\documentclass[lettersize,journal]{IEEEtran}
\usepackage{amsmath,amsfonts}

\usepackage{algorithm}                    
\usepackage{algorithmicx}                  
\usepackage[noend]{algpseudocode}

\usepackage{array}
\usepackage[caption=false,font=normalsize,labelfont=sf,textfont=sf]{subfig}
\usepackage{textcomp}
\usepackage{stfloats}
\usepackage{url}
\usepackage{verbatim}
\usepackage{graphicx}
\usepackage{multirow}
\usepackage[usenames,dvipsnames,table,xcdraw]{xcolor}  
\usepackage{hyperref}

\newcommand{\dof}{{\sc dof}}

\newcommand{\cspace}{\ensuremath{\mathcal{C}_{space}}}
\newcommand{\cspaces}{\ensuremath{\mathcal{C}_{spaces}}}

\newcommand{\tsh}{\ensuremath{\mathcal{H}_\mathcal{T}}}
\newcommand{\tsv}{\ensuremath{\mathcal{V}_\mathcal{T}}}
\newcommand{\tse}{\ensuremath{\mathcal{E}_\mathcal{T}}}
\newcommand{\moh}{\ensuremath{\mathcal{H}_\mathcal{M}}}
\newcommand{\mov}{\ensuremath{\mathcal{V}_\mathcal{M}}}
\newcommand{\moe}{\ensuremath{\mathcal{E}_\mathcal{M}}}
\newcommand{\teh}{\ensuremath{\mathcal{H}_{TE}}}
\newcommand{\tev}{\ensuremath{\mathcal{V}_{TE}}}
\newcommand{\tee}{\ensuremath{\mathcal{E}_{TE}}}

\def\BibTeX{{\rm B\kern-.05em{\sc i\kern-.025em b}\kern-.08em
    T\kern-.1667em\lower.7ex\hbox{E}\kern-.125emX}}
\usepackage{balance}
\begin{document}
\title{Hypergraph-based Multi-Robot Task and Motion Planning}

\author{James Motes, Tan Chen, Timothy Bretl, Marco Morales, Nancy M. Amato
\thanks{Manuscript created April, 2023.\\
             James Motes (corresponding author), Marco Morales, and Nancy M. Amato are with Parasol Lab, Department of Computer Science, University of Illinois at Urbana-Champaign (email: \{jmotes2, moralesa, namato\}@illinois.edu).\\
             Tan Chen is with the Department of Electrical and Computer Engineering at Michigan Technological University (email: tanchen@mtu.edu). \\
             Timothy Bretl is with the Department of Aerospace Engineering, University of Illinois at Urbana-Champaign (email: tbretl@illinois.edu).\\
             This work was supported in part by Foxconn Interconnect Technology (FIT) and the Center for Networked Intelligent Components and Environments (C-NICE) at UIUC.
             }}


\maketitle

\begin{abstract}
We present a multi-robot task and motion planning method that, when applied to the rearrangement of objects by manipulators, 
results in solution times up to three orders of magnitude faster than existing methods \textcolor{black}{and 
successfully plans for problems with up to twenty objects, more than three times as many objects as comparable methods.}
We achieve this improvement by decomposing the planning space to consider manipulators alone, objects, and manipulators holding objects. 
We represent this decomposition with a hypergraph where vertices are decomposed \textcolor{black}{elements} of the planning spaces and hyperarcs are transitions between \textcolor{black}{elements}. 
Existing methods use graph-based representations where vertices are full composite spaces and edges are transitions between these. 
Using the hypergraph reduces the representation size of the planning space—for multi-manipulator object rearrangement, the number of hypergraph vertices scales linearly with the number of either robots or objects, while the number of hyperarcs scales quadratically with the number of robots and linearly with the number of objects. 
In contrast, the number of vertices and edges in graph-based representations scales exponentially in the number of robots and objects. 
We show that similar gains can be achieved for other multi-robot task and motion planning problems.
\end{abstract}

\section{Introduction}
\IEEEPARstart{T}{he} use of autonomous robotic systems is rapidly increasing.
This can be seen in warehouse management, manufacturing, household chores, health care, etc..
In many of these settings, multi-robot systems are leveraged to complete tasks single robots cannot through inter-robot cooperation and increase throughput by working in parallel.
These improvements in effectiveness often require complex coordination for both task and motion planning.

Unfortunately, the planning space for these problems is very large.
Even for a single robot, it is intractable in the general case to represent explicitly~\cite{ss-otpmpiigtfctporam-83,c-crmp-88}, and it grows exponentially for the composite space of multi-robot systems.
Task and motion planning adds additional dimensions for the tasks present.

\begin{figure}[t]
    \centering
    \includegraphics[width=1\linewidth]{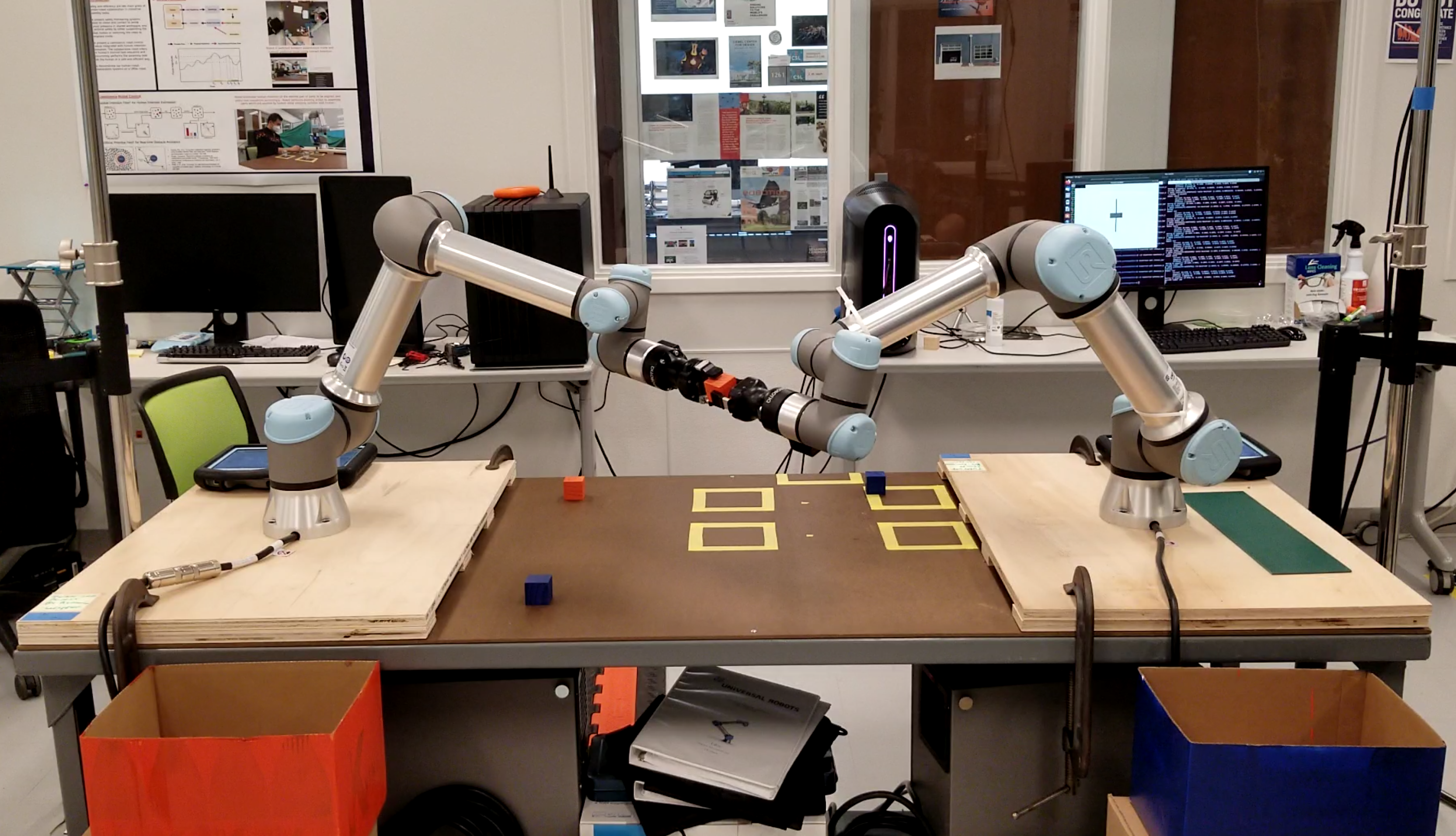}
    \caption{
        We demonstrate our proposed approach for the multi-manipulator sorting problems with a pair of UR5e manipulators. The robots must sort the blocks into bins of the corresponding color. This figure captures the moment in which the robots are performing a handoff. A full video of the sorting demo can be found at: \href{https://youtu.be/MSyeYXu0Xzs}{https://youtu.be/MSyeYXu0Xzs}.
    }
    \label{fig:physical_image}
\end{figure}

For simpler problems, \textcolor{black}{decoupling the planning space into individual robot state spaces} results in faster planning times.
However, decoupled approaches cannot handle problems which require high levels of coordination.
Thus, existing multi-robot task and motion planning approaches primarily plan in the composite space and cannot efficiently plan for large complex multi-robot systems.

Some problems, such as payload transportation, model tasks abstractly and often ignore the physical dimensions of the tasks during planning~\cite{mslta-mrtampwsd-20,bpssk-ostaapffmrap-20,hat-aoatstctaapfp-19}.
Other problems, especially those involving object manipulation, must consider physical constraints~\cite{2014-wafr-glk, hl-itaprmmpdwmimpq-09,db-praafpmm-15,sb-amtampfpoiovh-19,sb-smrgbmgwcc-20}.
Here the coordination required limits the use of decoupled methods, and the planning space grows too large to apply composite methods.

Hybrid search techniques that balance the strengths and weaknesses of coupled and decoupled approaches have shown promise in recent multi-robot motion planning methods~\cite{smsa-rmmpucs-21,cukk-oabmmp-19}.
However, before our work, hybrid search techniques were not used for problems requiring more complex coordination such as multi-robot task and motion planning.

In this paper, we present the \textit{Decomposable State Space Hypergraph} (DaSH) method, a general \textcolor{black}{multi-robot planning framework based on a hypergraph representation that can both incorporate existing composite and decoupled multi-robot methods and enable new hybrid methods.
We present a novel hybrid search technique based on the DaSH representation} which captures the coordination and complexity in multi-robot task and motion planning problems while scaling to larger numbers of robots and tasks.
\textcolor{black}{Additionally, our method successfully plans for higher ratios of tasks to robots (10:1) which existing methods such as~\cite{sb-smrgbmgwcc-20} struggle with.}

Our approach supports varying levels of (de)coupling of the planning space at different problem stages.
The degree of coupling should be tailored to the problem and its corresponding planning space and can change to fit different problem stages.
We support this with hybrid composition representations for several planning spaces which capture varying degrees of (de)coupled spaces depending on the coordination required.

Take, for example, a multi-manipulator problem involving a handoff between two robots as depicted in Figure~\ref{fig:physical_image}.
Different levels of coordination are required at different stages of planning.
Robots operate mostly independently in their individual state spaces.
The joint space of a single robot and an object is important while picking/placing the object or while carrying the object,
and the joint space of both robots and an object is important while planning a handoff between the two robots.

\begin{figure}[t]
\centering
 \subfloat[2 Robots, 4 Objects\\Graph - Hypergraph]{
	\includegraphics[width=.8\linewidth]{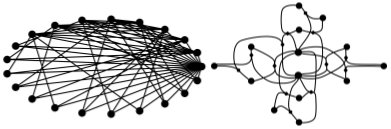}
	\label{fig:intuitive_graphs_fig}
}\\
\subfloat[Representation Sizes]{
\small
\begin{tabular}{|c|c|crr|}
\hline
Robots              & Objects             & \multicolumn{1}{c|}{Representation} & \multicolumn{1}{c|}{Vertices} & \multicolumn{1}{c|}{Transitions} \\ \hline
                    &                     & Hypergraph                          & 14                            & 24                               \\
                    & \multirow{-2}{*}{4} & \cellcolor[HTML]{EFEFEF}Graph       & \cellcolor[HTML]{EFEFEF}21    & \cellcolor[HTML]{EFEFEF}120      \\ \cline{2-2}
                    &                     & Hypergraph                          & 26                            & 48                               \\
\multirow{-4}{*}{2} & \multirow{-2}{*}{8} & \cellcolor[HTML]{EFEFEF}Graph       & \cellcolor[HTML]{EFEFEF}73    & \cellcolor[HTML]{EFEFEF}496      \\ \cline{1-2}
                    &                     & Hypergraph                          & 24                            & 80                               \\
                    & \multirow{-2}{*}{4} & \cellcolor[HTML]{EFEFEF}Graph       & \cellcolor[HTML]{EFEFEF}209   & \cellcolor[HTML]{EFEFEF}8672     \\ \cline{2-2}
                    &                     & Hypergraph                          & 44                            & 160                              \\
\multirow{-4}{*}{4} & \multirow{-2}{*}{8} & \cellcolor[HTML]{EFEFEF}Graph       & \cellcolor[HTML]{EFEFEF}3393  & \cellcolor[HTML]{EFEFEF}213184   \\ \hline
\end{tabular}
}
\caption{
The figures and table provide an illustration of the contrast in representation sizes for composite (graph) and decoupled/hyrbid (hypergraph) search spaces for small multi-manipulator problems.
The details of the problem constraints can be found in Section~\ref{section:representation_analysis}.
(a) The composite graph-based representation of the task space for 2 robots and 4 objects contains 21 vertices and 120/60 directed/bi-directional edges.
The decoupled hypergraph-based representation for the same problem contains 14 vertices and 24/12 directed/bi-directional hyperarcs. 
(b) The number of vertices and directed transitions (edges/hyperarcs) are given for increasing problem sizes of 2-4 robots and 4-8 objects. 
The reduction in size of the hypergraph representation is significant when increasing both robots and objects. 
The graph representation quickly becomes too large to use efficiently, while the hypergraph remains relatively small.
}\label{fig:intuitive_graphs}
\end{figure}

\textcolor{black}{
These decoupled and hybrid composition representations often remain sparse as the number of robots and tasks (or objects) increase.
In contrast, the size of pure composite representations scale exponentially with both the number of robots and tasks.
A small illustration of the size difference is shown in Fig.~\ref{fig:intuitive_graphs}.
As a result, existing methods which plan in the composite space are forced to use implicit representations of the composite space and engineer heuristics which make assumptions about the underlying problem structure.
Our decoupled and hybrid representations can be explicitly constructed and leveraged into powerful heuristics informed directly by the representation leading to significant performance gains.
}

Additionally, prior multi-manipulator methods such as~\cite{sb-smrgbmgwcc-20} struggle as the number of tasks (or objects) increases relative to the number of robots.
Our proposed representation flips the scaling paradigm where tasks (or objects) are no longer the limiting factor.
As a result, DaSH supports higher ratios of objects to robots which is appropriate for manipulation tasks where a few robots should handle many objects.

We demonstrate the ability of the DaSH method to generalize to many multi-robot task and motion planning problems including \textcolor{black}{multi-robot motion planning and multi-manipulator rearrangement.
For the rearrangement planning problem}, we illustrate how the number of vertices in the hypergraph scales linearly with both the number of robots and objects, while the number of hyperarcs scales quadratically with the number of robots and linearly with the number of objects (Fig.~\ref{fig:intuitive_graphs}) and demonstrate improved planning times on both physical (Fig.~\ref{fig:physical_image}) and virtual multi-robot systems (Fig.~\ref{fig:shelves_image}).

In summary, our contribution is:
\begin{itemize}
    \item \textcolor{black}{A generalized definition of the multi-robot planning space which generalizes composite, decoupled, and hybrid approaches.}
    \item A novel hypergraph representation for varying coupled/decoupled spaces in multi-robot planning problems.
    \item A general planning algorithm that exploits this hypergraph representation.
    \item The application of this approach to (i) the multi-robot motion planning space (MRMP-DaSH) and to (ii) the multi-manipulator planning space (MM-DaSH).
    \item A theoretical analysis of the size of the proposed hypergraph representation compared to traditional graph representations for multi-manipulator \textcolor{black}{rearrangement} problems.
    \item An experimental evaluation of the planning algorithm applied \textcolor{black}{to the multi-manipulator rearrangement problem}.
\end{itemize}

\section{Background and Related Work}
In this section, we give an overview of the background and related work for multi-robot task and motion planning.

\subsection{Motion Planning}
The \textit{degrees of freedom} ({\dof}s) of a robot fully parameterize its position. 
They may contain the robot's pose, orientation, joint angles, etc..
A specification of {\dof} values for a robot defines a \textit{configuration}.
Motion planning considers a continuous state space comprised of the set of all robot configurations known as \textit{configuration space} (\cspace)~\cite{lw-apcfpapo-79}.

The motion planning problem is to find a continuous path from a start location to a goal location through the subset of $\cspace$ consisting of valid configurations called \textit{free space}.
$\cspace$ is usually intractable to represent explicitly~\cite{ss-otpmpiigtfctporam-83,c-crmp-88}. 
To handle the complexity of planning in {\cspace}, sampling-based motion planners such as the Probabilistic Roadmap Method (PRM)~\cite{kslo-prpp-96} attempt to create a discretized approximation of the connectivity of the $\cspace$ known as a roadmap.
A roadmap is a graph in which vertices are individual configurations and edges represent transitions between a pair of configurations. 
Paths are found by searching over this roadmap.

\subsection{Multi-robot Motion Planning}
\label{section:related_work_mrmp}

Multi-robot motion planning considers the composite $\cspace$ of the system. 
This composite space is defined as the Cartesian product of the $\cspace$ of each individual robot.
This space grows exponentially with the number of robots.

There are three main approaches to planning within this space.
\textit{Coupled} planners plan directly within the composite space~\cite{sl-uppccdpmrs-2002,ssh-faniaehdrfeoirimm-16}. 
They struggle to plan for large numbers of robots and are often slow, but they are able to provide probabilistic completeness. 
\textit{Decoupled} methods consider individual robot $\cspace$ to find a set of paths~\cite{sl-uppccdpmrs-2002,si-mmpbic-06}.
They are typically faster but usually lack completeness and optimality guarantees.
\textit{Hybrid} methods such as CBS-MP~\cite{smsa-rmmpucs-21} seek to leverage the strengths of both composite and decoupled planners.
They often iteratively plan within individual robot $\cspaces$ (or the composite space of a subgroup of the robots) while reconciling plans against each other.
Hybrid methods usually \textcolor{black}{achieve planning times similar to} decoupled methods while offering the completeness and optimality guarantees of composite methods.

\subsection{Task and Motion Planning}
When we consider planning problems with objects that can be moved by robots, we enter the domain of task and motion planning~\cite{gchkskl-itamp-21}.
If we model the state of the entire world (robots and objects) as a configuration space $\mathcal{W}$, we end up with a highly under actuated system. 
We only have direct control over the robot {\dof}s, 
while the {\dof}s corresponding to any object pose can only be changed indirectly through the robot acting upon them.
Additionally, the actions the robot takes affect the feasibility of motions.
For example, a valid motion for a robot not holding an object may no longer be valid if the robot is grasping an object, and the current placement of the objects affects the feasibility of motions. 

One approach is to treat each combination of the robot and held object as a unique robot and each placement of objects as a new environment.
This is sometimes referred to as multi-modal motion planning~\cite{hl-mmmpnes-10,hn-rmmpfahrmat-11,bkl-ahatmwda-13} although this language is less intuitive than task and motion planning and can be confused for motion planning with different modes of locomotion.

Instead, we define a \textcolor{black}{\textit{task space} $\mathcal{T}$, where element $T_i\in\mathcal{T}$} defines which object the robot is holding (or none), the grasp constraints for the held object, and valid poses for all ungrasped objects.
This defines a new configuration space $\mathcal{W}_i$ for the robot (possibly while grasping an object).
\textcolor{black}{Details of this are provided in Section~\ref{section:prob_def}.}

A system can switch between \textcolor{black}{task space elements} by applying an action that changes the constraints of valid motion. 
For example, performing a pick action on an object removes the constraint that the object is on a stable surface and adds the constraint that the object is in a stable grasp for the robot.

A system can only switch between task space elements at configurations \textcolor{black}{that are valid within the two task space elements that it is switching between}.
In practice, these are often stable grasps of objects that are sitting at stable poses on some surface in the environment or handoffs at the intersection of stable grasp poses for a set of robots~\cite{db-praafpmm-15}.
These configurations satisfy \textcolor{black}{the constraints of the task space element defining} the robot grasping the object and the task space element where the object is located at that stable pose.
Depending on the direction of the switch, these configurations can represent the moment of a pick or place action.
Finding these transition configurations is considered the most challenging part of multi-modal planning~\cite{gchkskl-itamp-21}. 


When considering classic task and motion planning problems, it is often best to view them as a hybrid discrete-continuous search problems~\cite{gchkskl-itamp-21}.
A solution consists of a finite and discrete sequence of task space elements (eg: which object to grasp) with
continuous constraint parameters (poses and grasps of objects), 
and continuous motion paths within the configuration space $W_i$ of each task state $T_i$ to a configuration in the intersection with the subsequent task space element's configuration space.

\subsection{Multi-robot Task and Motion Planning}

In multi-robot systems, a task space element consists of the same types of constraints (object poses and grasp constraints in manipulation problems) as in single robot systems.
However, in this domain, there is a combinatorial expansion in the number of task space elements available to the system as both the number of robots and objects grow~\cite{s-tpomemmtampwog-20}.

The three classes of multi-robot motion planning (coupled, decoupled, and hybrid) apply here as well.
Most task and motion planning problems require too much coordination for decoupled methods to be useful.
Although some of our prior work has explored hybrid search techniques in the multi-robot task and motion planning domain, the approach does not extend to manipulation problems~\cite{mslta-mrtampwsd-20}.
The rest of this section discusses coupled approaches.

\begin{figure*}[t]
\centering
 \subfloat[Stable Pose - Start]{
	\includegraphics[width=.2\linewidth]{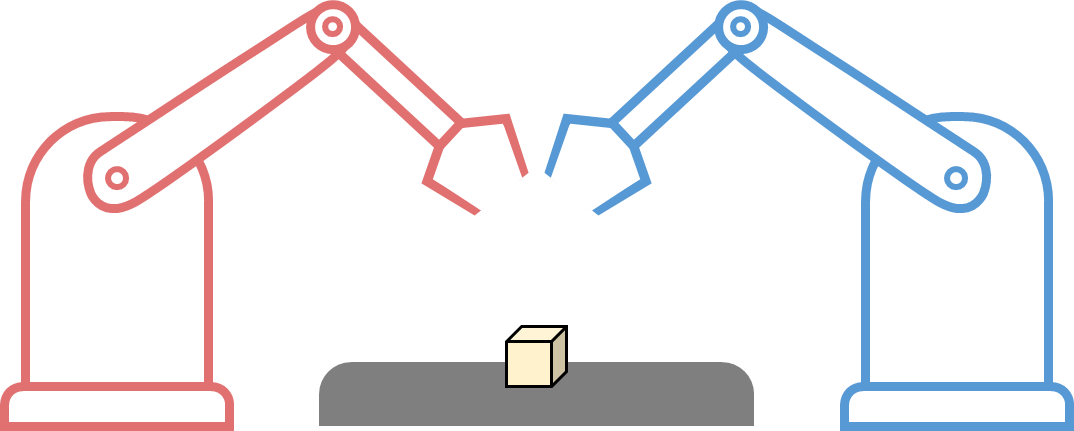}
	\label{fig:multi-kg-stable-pose}
}\hspace{.25in}
 \subfloat[Grasp 1]{
	\includegraphics[width=.2\linewidth]{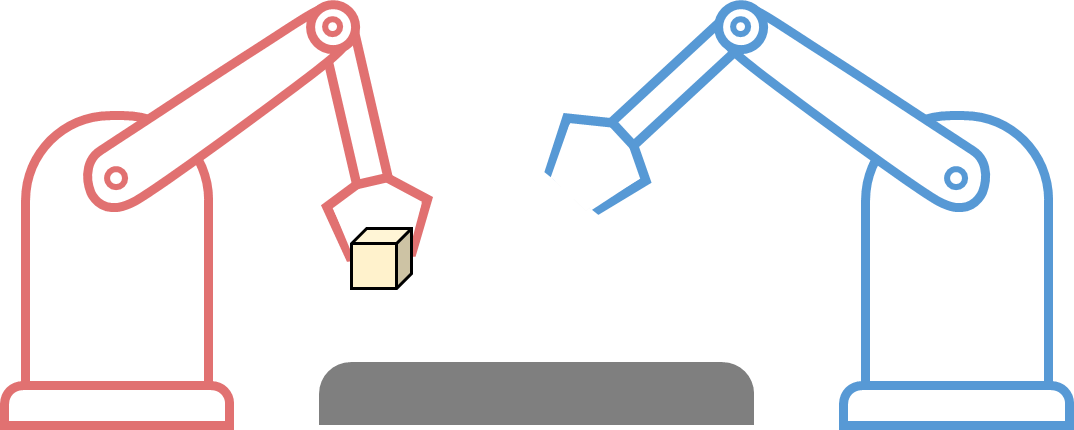}
	\label{fig:multi-kg-grasp1}
}\hspace{.25in}
 \subfloat[Grasp 2]{
	\includegraphics[width=.2\linewidth]{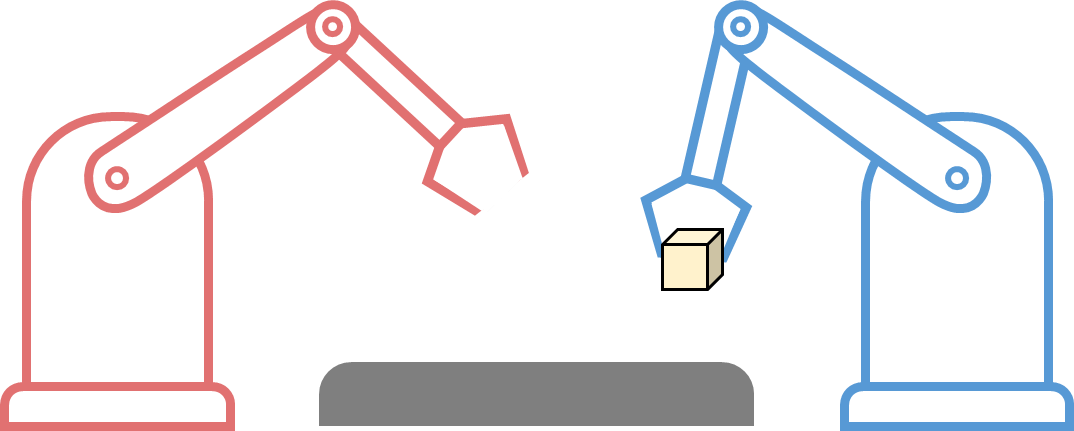}
	\label{fig:multi-kg-grasp2}
}\hspace{.25in}
\subfloat[Stable Pose - Goal]{
	\includegraphics[width=.2\linewidth]{images/manip_stable_pose_full.png}
	\label{fig:multi-kg-stable-pose2}
}
\caption{
	a) The figure shows the starting task space element of the system where the object is at rest at a stable pose on the table.
	b) A \textit{pick} action results in a switch to the middle left task space element where the red robot is grasping the object.
	c) A \textit{handoff} action creates another switch transferring the object from the red robot to the blue robot.
	d) Finally, a \textit{place} action returns the system to a stable pose task space element.
}\label{fig:multi-kg}
\end{figure*}

Previous work in the area has had two main focuses to handle the space complexity: developing representations and developing heuristics.
The authors in~\cite{ufm-aitamptfm-19} use a shared manipulator workspace to build a shared space graph to reason over multi-robot cooperation and adapt a path planning heuristic for multi-manipulator planning.
Assumptions about reachability and the number of robots required to securely grasp an object is used in~\cite{db-praafpmm-15} to build a condensed graph representation.
This representation is integrated in later work with a multi-robot motion planning method~\cite{ssdhb-dsaiammp-20} to create a full multi-robot multi-modal planner~\cite{sb-amtampfpoiovh-19}.
This planner introduces a high-level object mode graph to represent valid transitions between pick, place, and handoff configurations.
This graph is used as a heuristic to guide the multi-modal motion planning through sequences of modes that form a valid solution.
All three of these methods~\cite{ufm-aitamptfm-19,db-praafpmm-15,sb-amtampfpoiovh-19} only plan for a single object.

In~\cite{sb-smrgbmgwcc-20}, the work in~\cite{sb-amtampfpoiovh-19} is expanded to account for multiple objects.
The single object mode graph in~\cite{sb-amtampfpoiovh-19} is replaced with an object centric mode graph where vertices correspond to manipulators and stable surfaces.
Instead of searching over object (or task state) switches for a single object, a multi-agent pathfinding (MAPF) technique is proposed to find non-conflicting individual object mode sequences over the available manipulators and stable surfaces.
The transfer of multiple objects at once complicates the reasoning over both the sequence of modes and the motion planning.
The authors simplify the problem by considering synchronized actions for the set of manipulators in the problem.
This allows a single step in the MAPF solution over the object centric mode graph to both indicate a set of actions for the robots to perform and define a more constrained multi-robot motion planning problem.

Using MAPF over the object centric mode graph as a heuristic, the Synchronized Multi-Arm Rearrangement (SMART) is able to efficiently plan for up to 9 robots~\cite{sb-smrgbmgwcc-20}.
The heuristic is formulated as a MAPF problem over the object centric mode graph.
The method is biased to greedily move the object centric mode state forward along the corresponding MAPF solution.
When the MAPF solution is simple and the paths for individual objects are not likely to use the same robots at the same time, this greedy heuristic performs exceptionally well.
However, when the paths for objects are likely to conflict, or the number of objects increases relative to the number of robots (increasing the density of the MAPF problem), this heuristic becomes limiting.
This is reflected in~\cite{sb-smrgbmgwcc-20} as no result is shown for more than 4 objects.
We directly compare our method against SMART~\cite{sb-smrgbmgwcc-20} in later sections.

\textcolor{black}{
Recent work such as ~\cite{pwsk-agtampffmm-21} have addressed more complicated manipulation tasks such as tower stacking and obstructing obstacles though they still struggle to scale the number of objects relative to the number of robots with the highest ratio coming with 6 objects for 2 robots.
The evaluation of the method's ability to handle increasing numbers of robots only considers very simple task problems where each robot has a very clear correct role~\cite{pwsk-agtampffmm-21}.
We show the ability to solve similar problems in Section~\ref{section:experiments} and obtain higher ratios of objects to robots in our more general experiments.
}

\textcolor{black}{
All of these methods focus on a single short term task involving a small set of obstacles.
In contrast,~\cite{hodot-lmrpfca-22} presents a long-horizon multi-robot task and motion planning construction method which decomposes problems with up to 113 objects into subproblems each focused around a single object.
Methods such as the one presented here and~\cite{pwsk-agtampffmm-21} can be integrated into the long-horizon framework of~\cite{hodot-lmrpfca-22} to address more complicated subtasks within long-horizon problems.
We do not address this integration in this work.
}

\subsection{Directed Hypergraphs and Hyperpaths}
\label{section:directed_hypergraphs}
A hypergraph $\mathcal{H}=(\mathcal{V},\mathcal{E})$ is a generalization of a graph, where $\mathcal{V}=\{v_0,v_1,...,v_n\}$ is the set of vertices, and $\mathcal{E}=\{E_0,E_1,...E_m\}$, with $E_i\subset\mathcal{V}$ for $i\in [0,m]$, is the set of hyperedges.
Unlike edges in a regular graph, a hyperedge $E\in\mathcal{E}$ is not restricted to pairs of vertices.
In this work, we consider directed hypergraphs as described in~\cite{glpn-dhaa-93}.

\subsubsection{Directed Hypergraph}
\begin{figure}[h]
\centering
 \subfloat[Directed Hyperarc]{
	\includegraphics[width=.45\linewidth]{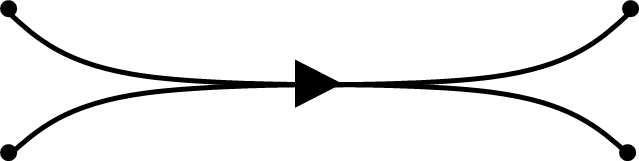}
	\label{fig:directed_hyperarc}
}
 \subfloat[Bi-Directional Hyperarc]{
	\includegraphics[width=.45\linewidth]{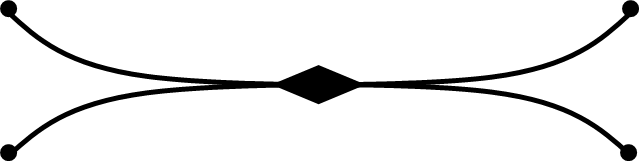}
	\label{fig:bidirected_hyperarc}
}
\caption{
	a) A directed hyperarc represents a transition from the vertices in the \textit{tail set} on the left to the vertices in the \textit{head set} on the right.
	b) A bi-directional hyperarc indicates that the transition can happen in either direction. It is essentially a pair of directed hyperarcs with opposite tail and head sets.
}\label{fig:hyperarcs}
\end{figure}

\begin{figure*}[h]
\centering
 \subfloat[Directed Hypergraph]{
	\includegraphics[width=.2\linewidth]{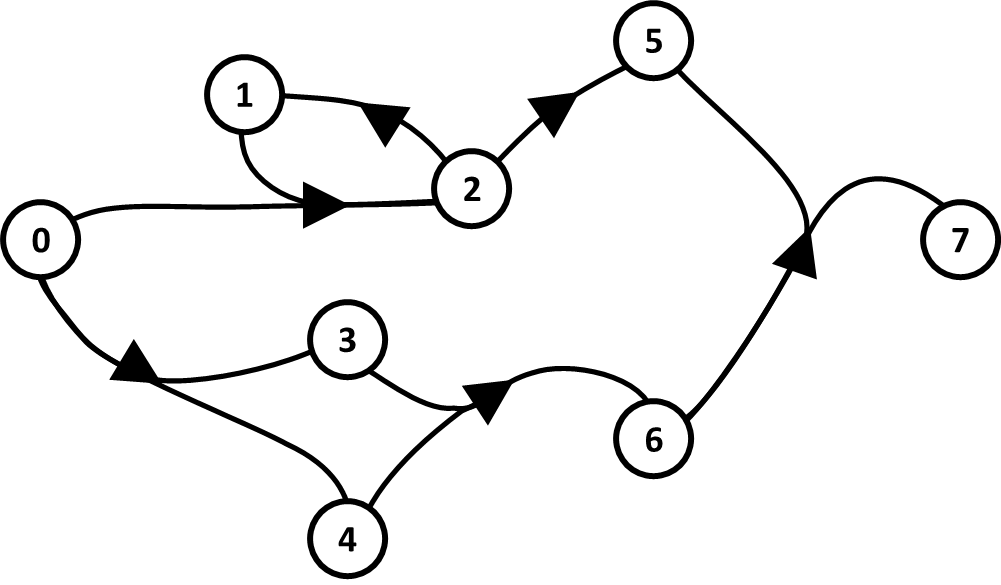}
	\label{fig:directed_hypergraph}
}\hspace{.6in}
 \subfloat[Simple Path]{
	\includegraphics[width=.2\linewidth]{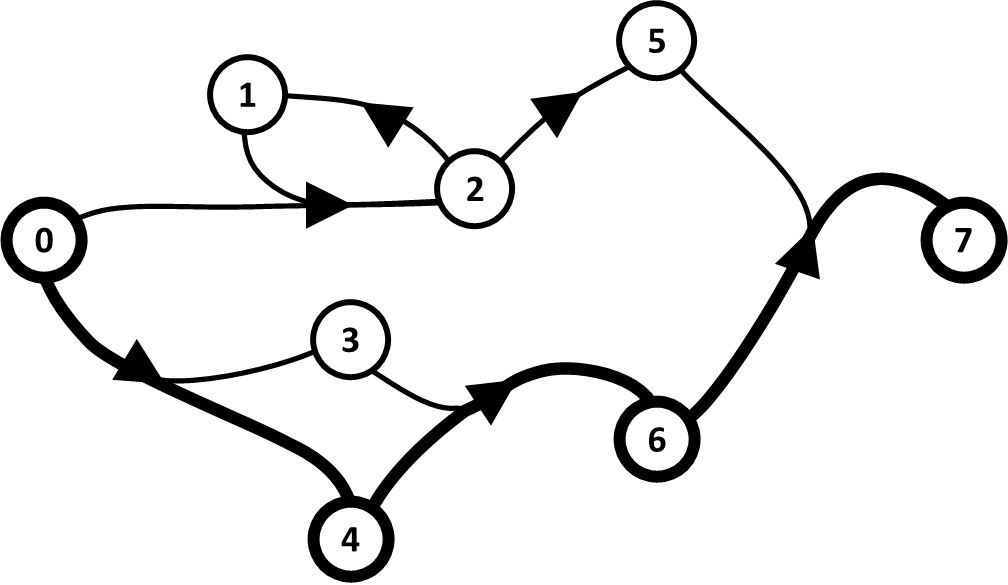}
	\label{fig:directed_simple_path}
}\hspace{.6in}
 \subfloat[Hyperpath Path]{
	\includegraphics[width=.2\linewidth]{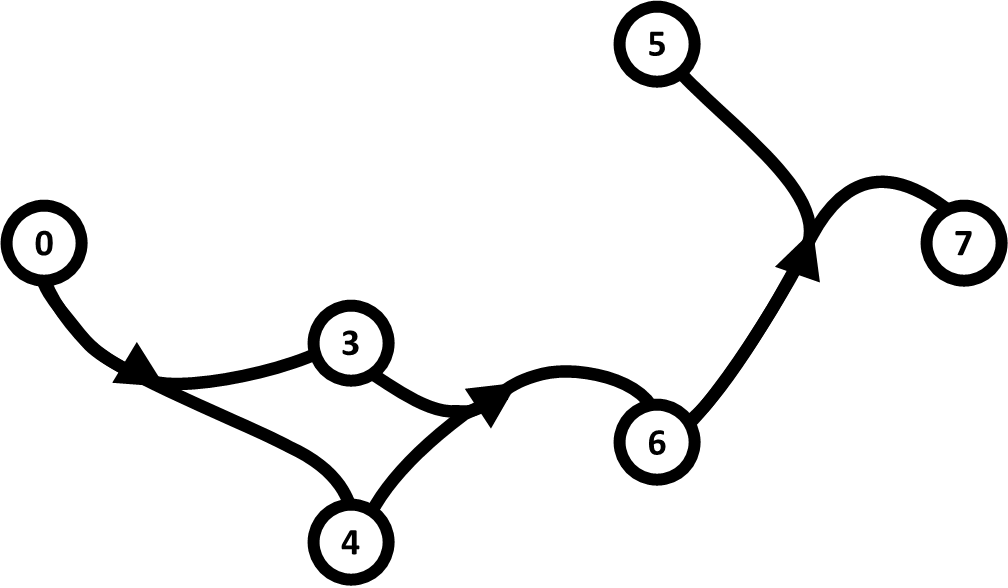}
	\label{fig:directed_hyperpath}
}
\caption{
	a) A directed hypergraph consists of vertices and directed hyperarcs.
	(b) A simple path in a directed hypergraph from $v_0$ to $v_7$. No hyperarc is used more than once.
	(c) A directed hyperpath from $v_0$ to $v_7$. Every vertex incident a hyperarc in the hyperpath is included. All vertices are connected to the source $v_0$ by a cycle-free simple path.
}\label{fig:hypergraph}
\end{figure*}

A directed hypergraph has directed hyperedges, or \textit{hyperarcs}, so that a hyperarc $E_i$ has both a head set $\texttt{Head}(E_i)\subseteq\mathcal{V}$ and a tail set of vertices $\texttt{Tail}(E_i)\subseteq\mathcal{V}$.
A visual depiction of directed hyperarcs is given in Figure~\ref{fig:hyperarcs},
and a depiction of a directed hypergraph is given in Figure~\ref{fig:directed_hypergraph}.
In this paper, all hypergraphs discussed will be directed hypergraphs.

\subsubsection{Directed Hyperpath}
\label{section:directed-hyperpath}
We use directed hyperpaths as defined in~\cite{glpn-dhaa-93}.
A \textit{path} in a hypergraph is a sequence of vertices and hyperarcs $v_0,E_0,v_1,E_1,...,v_{final}$ such that $v_i\in \texttt{Tail}(E_{i})$ and $v_{i+1}\in \texttt{Head}(E_{i})$ for all vertices and edges in the path (Fig.~\ref{fig:directed_simple_path}).
A path is \textit{simple} if all hyperarcs are used at most once.

A \textit{hyperpath} $\Pi_{st}$ is a minimal hypergraph $\mathcal{H}_{\Pi}=(\mathcal{V}_{\Pi},\mathcal{E}_{\Pi})$ such that:
\begin{align}
    \label{eq:hyperpath_arcs}
    &\mathcal{E}_{\Pi}\subseteq\mathcal{E}\\
    \label{eq:hyperpath_vertices}
    &s,t\in\mathcal{V}_{\Pi}=\bigcup_{E_i\in\mathcal{E}_{\Pi}}E_i\subseteq\mathcal{V}\\
    \label{eq:hyperpath_connected}
    \begin{split}
        &x\in\mathcal{V}_{\Pi}\Rightarrow x\text{ is connected to $s$ in $\Pi_{st}$} \\
        & \text{ through a cycle-free simple path.}
    \end{split}
\end{align}

The set of hyperarcs $\mathcal{E}_{\Pi}$ in the hyperpath $\Pi_{st}$ must be in the original hypergraph $\mathcal{H}$ (\ref{eq:hyperpath_arcs}).
The set of vertices  $\mathcal{V}_{\Pi}$ consists of all vertices incident to a hyperarc $E_i\in\mathcal{E}_{\Pi}$, and the source $s$ and target $t$ must be included in $\mathcal{V}_{\Pi}$ (\ref{eq:hyperpath_vertices}).
Every vertex must be connected to the source $s$ in the hyperpath through a cycle-free simple path (\ref{eq:hyperpath_connected}).
Thus the hyperarc from $\{v_0,v_1\}$ to $\{v_2\}$ in Figure~\ref{fig:hypergraph} cannot be included in a hyperpath from source $v_0$ as $v_1$ is not connected to $v_0$ via a cycle-free simple path.

\section{Problem Definition}
\label{section:prob_def}
In this section, we define the multi-robot task and motion planning problem and the corresponding planning space for a set of moveable bodies $\mathcal{B}$.
\textcolor{black}{
Our definition generalizes the coupled, decoupled, and hybrid representations used in multi-robot planning.
In Sections~\ref{section:mrmp} and~\ref{section:multi-manip}, we show how previous multi-robot motion planning (MRMP) and multi-robot task and motion planning (MR-TMP) methods fit into this definition, and how it enables the hybrid MR-TMP approach presented in Section~\ref{section:method}.
In the following paragraphs,}
we explain this definition in the context of \textcolor{black}{multi-robot motion planning and multi-manipulator rearrangement planning.} 

\subsection{Planning Space}
\textcolor{black}{
We consider a task space $\mathcal{T}$ consisting of elements $T_i=(B_i,\mathcal{W}_i,C_i)$ where $B_i\subseteq\mathcal{B}$ is a subset of the moveable bodies, $\mathcal{W}_i$ is the $\cspace$ for $B_i$ , and $C_i$ is a set of constraints which define validity in $\mathcal{W}_i$.
Constraints in $C_i$ may only apply to the moveable bodies in $B_i$, thus the validity of a configuration $w\in\mathcal{W}_i$ cannot depend on a moveable body $b\notin B_i$.
}

\textcolor{black}{
Task space elements can be further decomposed into additional sets of elements or combined with other elements.
We denote the decomposition of a task element $T_i=(B_i,\mathcal{W}_i,C_i)$ into a set of task space elements as $D({T}_i)\subseteq\mathcal{T}$.
A task space element $T_j=(B_j,\mathcal{W}_j,C_j)$ belongs to $D({T}_i)$ if $B_j\subseteq B_i$, $\mathcal{W}_j$ is a subspace of $\mathcal{W}_i$, and $C_j\subseteq C_i$.
This resembles a power set though $\mathcal{W}_j$ is a subspace of $\mathcal{W}_i$ and the rules on constraints in $C_j$ only applying to robots in $B_j$ etc..}

\textcolor{black}{
We denote the combination of task space elements $T_j=(B_j,\mathcal{W}_j,C_j),T_k=(B_k,\mathcal{W}_k,C_k)$ to create $T_i=(B_i,\mathcal{W}_i,C_i)$ as $T_j+T_k=T_i$.
In the resulting element $T_i$, $B_i=B_j\cup B_k$, and $W_i$ is the corresponding $\cspace$.
We assume that no moveable bodies may ever be in collision with each other, so $C_i=C_j\cup C_k\cup\{$no collision between $b_p,b_q\in B_i\}$.
}

\textcolor{black}{
Each multi-robot task and motion planning (MR-TMP) problem has a set of \textit{admissible} task space elements $\mathcal{T}^*\subseteq\mathcal{T}$ such that for any $T_i=(B_i,\mathcal{W}_i,C_i)\in\mathcal{T}^*$, the moveable bodies $B_i=\mathcal{B}$ and constraints $C_i$ define an obtainable set of configurations in $\mathcal{W}_i$ for the system (i.e. the system designer considers these acceptable configurations).
}

\textcolor{black}{
The multi-robot motion planning (MRMP) problem for a group of robots $\mathcal{B}$ is often defined to consider the composite configuration space $\mathcal{C}_0\times...\times\mathcal{C}_{|\mathcal{B}|-1}$ such that $\mathcal{C}_i$ is the configuration space for robot $B_i\in\mathcal{B}$.
A configuration is valid if no robot is in collision with any obstacle in the environment and no two robots are in collision with each other.
A MRMP solution consists of a continuous valid path from some start point $w_\texttt{start}$ to some goal point $w_\texttt{goal}$, where $w_\texttt{start},w_\texttt{goal}\in\mathcal{C}_0\times...\times\mathcal{C}_{|\mathcal{B}|-1}$.
}

\textcolor{black}{
The task space $\mathcal{T}_\texttt{MRMP}$ for the MRMP problem contains a single \textit{admissible} element In the MRMP problem, $\mathcal{T}^*=\{T_\texttt{COMPLETE}=(B_\texttt{COMPLETE},\mathcal{W}_\texttt{COMPLETE},C_\texttt{COMPLETE})\}$ such that $C_\texttt{COMPLETE}$ contains all the MRMP constraints, $B_\texttt{COMPLETE}=\mathcal{B}$, and $\mathcal{W}_\texttt{COMPLETE}=\mathcal{C}_0\times...\times\mathcal{C}_{|\mathcal{B}|-1}$.
Composite MRMP methods plan motions directly within $T_\texttt{COMPLETE}$.
}

\begin{figure}
    \centering
    \includegraphics[width=\linewidth]{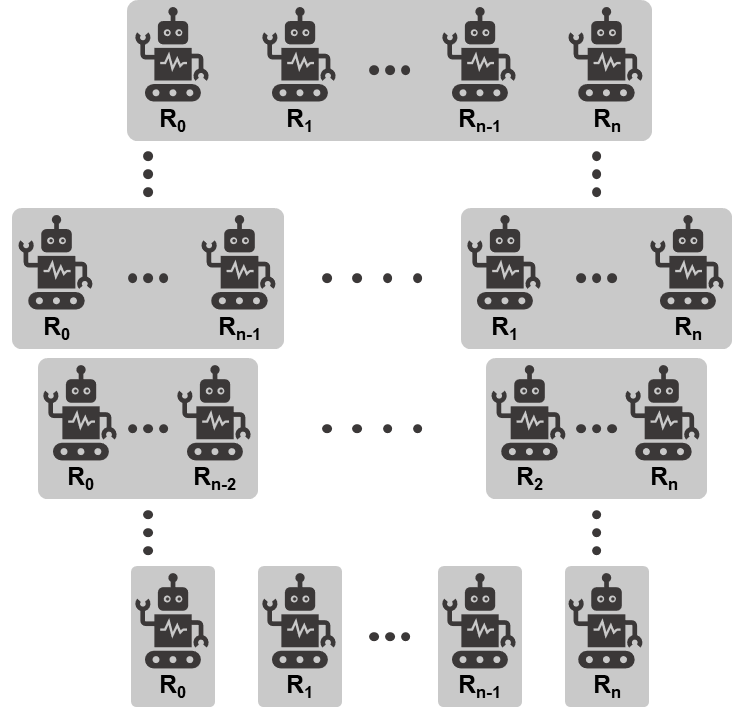}
    \caption{This figure depicts the task space for multi-robot motion planning. The top row indicates the admissible element containing all robots and collision constraints. This is the space composite methods plan in. The last row represents a the set fully decoupled elements, each containing a single robot. All other combinations of robots with their collision constraints also exist in the task space (in the intermediate rows).}
    \label{fig:mrmp_decomp}
\end{figure}

\textcolor{black}{
The full set of elements in $\mathcal{T}_\texttt{MRMP}$ can be obtained from the decomposition of $T_\texttt{COMPLETE}$ such that $\mathcal{T}_\texttt{MRMP}= D(T_\texttt{COMPLETE})$ as show in Fig.~\ref{fig:mrmp_decomp}.
Fully decoupled MRMP methods which plan motion separately for each robot $b_j\in\mathcal{B}$ plan within task elements $T_j=(B_j,\mathcal{W}_j,C_j)$ where $B_j=\{b_j\}$, $\mathcal{W}_j=\mathcal{C}_j$, and $C_j$ defines validity for $b_j$ independent of other robots.
Hybrid MRMP methods move between different levels of task space element decomposition and combination.
Both decoupled and hybrid MRMP methods construct a solution in $T_\texttt{COMPLETE}$ through combining their solutions in other elements.
This often requires the re-enforcement of the interrobot collision constraints relaxed when decomposing task space elements.
}

\textcolor{black}{
More general MR-TMP problems may have several admissible task space elements.
For example, an object manipulation problem may have an admissible element for every combination of robot-object grasps.
Furthermore, the start and goal of a problem may belong to different task space elements.
A valid motion path from the start to the goal must then \textit{transition} between task space elements.
}

\textcolor{black}{
A path may transition between a pair of elements $T_i,T_j$ at a transition configuration $w$ which is valid in $T_i+T_j$ so long as $T_i,T_j$ contain the same moveable bodies $B_i=B_j$.
A MR-TMP solution thus consists of both a sequence of elements in $\mathcal{T}^*$ and a valid continuous motion path through them.
}

\textcolor{black}{
In object manipulation problems, $\mathcal{T}^*\subseteq\mathcal{T}_\texttt{MANIP}$ contains elements which denote which robot is grasping which object.
These often resemble the task space elements depicted in Fig.~\ref{fig:multi-kg}).
Most existing multi-manipulator planning methods are composite approaches which plan only in task space elements in $\mathcal{T}^*$.
Transitions between task space elements (e.g. mode switches in~\cite{gchkskl-itamp-21,sb-smrgbmgwcc-20}) correspond to pick, place, and handoff actions.
Solutions consist of a sequence of these transitions between task space elements in $\mathcal{T}^*$ and valid paths within each task space element between transition configurations.
}

\textcolor{black}{
True decoupled approaches which plan with elements in the decompositions of admissible elements are likely to fail to find a plan for manipulation problems due to the transition requirement of two elements $T_i,T_j$ containing the same moveable bodies $B_i=B_j$.
For example, elements $T_i=(B_i,\mathcal{W}_i,C_i),T_j=(B_j,\mathcal{W}_j,C_j)$ where $B_i$ contains only robot $r_i$ and $B_j$ contains only object $o_j$, cannot transition to an element $T_k=(B_k,\mathcal{W}_k,C_k)$ where $B_k=\{r_i,o_j\}$ in a decoupled approach thus $r_i$ cannot grasp $o_j$.
Considering an element which contains both $r_i$ and $o_j$ before the grasp constraint is applied instead of $T_i,T_j$ addresses this issue, however, when $r_i$ needs to manipulate another object or $o_j$ needs to be manipulated by another robot the same issues arises.
This quickly forces decoupled approaches to only consider the admissible elements to find a feasible solution which is equivalent to composite approaches.
}

\begin{figure}
    \centering
    \includegraphics[width=.7\linewidth]{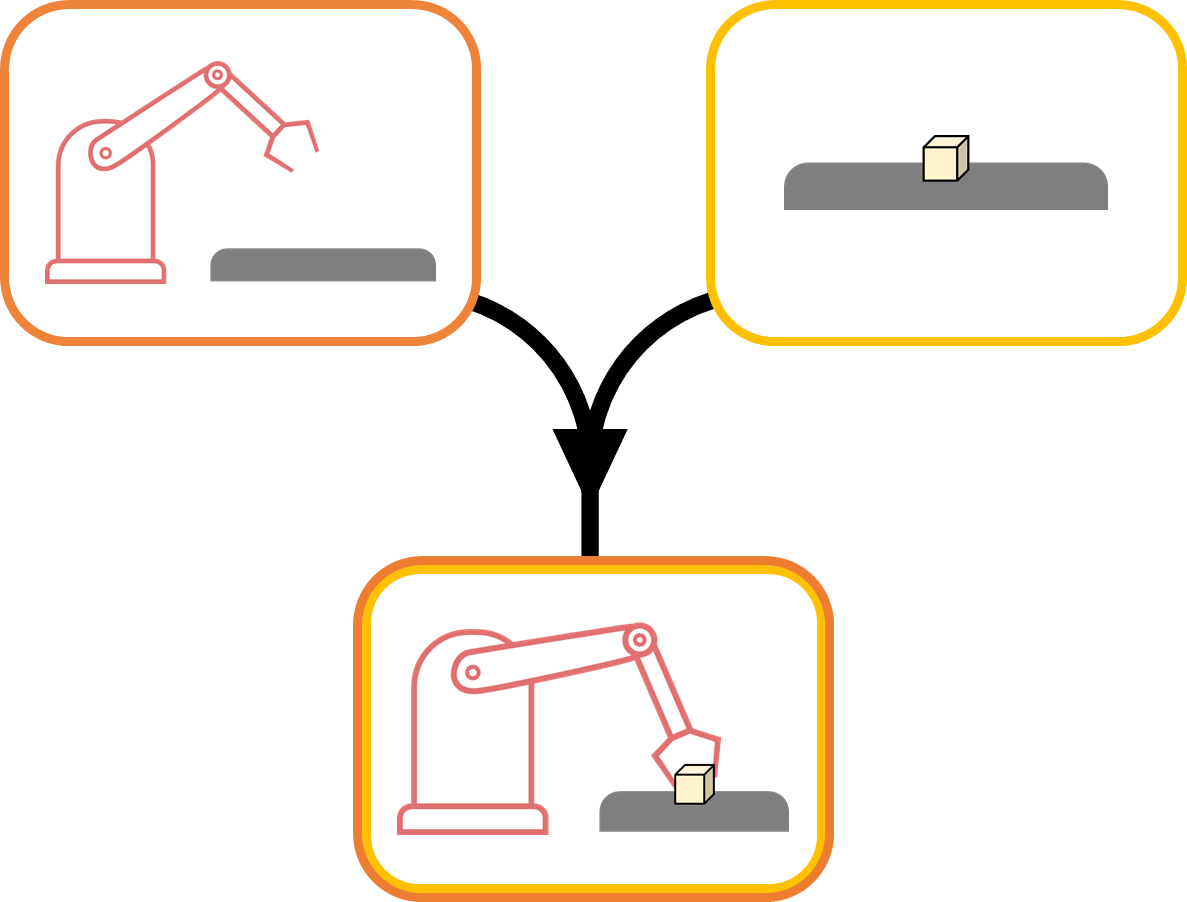}
    \caption{The hyperarc captures the grasp of an object. 
	      The transition requires both the task space element for the robot not holding an object, and the object not held and resting on a stable surface. 
	      These together form the tail set.
	      The transition moves the robot and object into a new task space element where the robot is now grasping the object.}
    \label{fig:transition-manip}
\end{figure}

\textcolor{black}{
However, hybrid methods can effectively plan with decoupled non-admissible elements. 
These approaches, which can move between different levels of task space element decomposition and combination,
can transition between $T_i+T_j=T_l$ and $T_k$ so long as $B_i\cup B_j=B_k$ and there exists a valid configuration in both $\mathcal{W}_l$ and $\mathcal{W}_k$ (Fig.~\ref{fig:transition-manip}).
Thus transitions can occur between sets of task space elements, so long as there exists a transition between the combination of elements on either side of the transition.
Just like in MRMP, hybrid methods must ensure that a solution constructed in elements decomposed from elements in $\mathcal{T}^*$ is a valid solution in the elements in $\mathcal{T}^*$.
This paper presents presents a hybrid approach to MR-TMP.
}

\section{Method}
\label{section:method}
We propose the \textit{Decomposable State Space Hypergraph} (DaSH) method, a general multi-robot task and motion planning method based on the problem formulation presented in Section~\ref{section:prob_def}.
We first present an overview of the method, followed by a detailed description of the representation and construction process.
We then define the task search and motion conflict resolution stage.
Finally, we discuss several variants of the method.

\subsection{Overview}
\label{section:method_overview}
Our approach consists of three stages: representation construction, task planning, and conflict resolution.

The DaSH method first builds the representation layers then queries them to generate task plans.
We introduce a hierarchy of hypergraph-based representation layers which capture increasing levels of information.
The highest layer contains the task space and transitions between task space elements.
The middle layer encodes motion feasibility within the constrained configuration spaces in the task space elements.
The final layer represents the search itself and includes the transition history of potential solutions.

Task plans are computed by finding a hyperpath through this final representation layer. 
This includes continuous motion paths within different task space elements.
Thus, a final conflict resolution stage ensures there are no inter-path collisions between different task space elements. 
A solution then consists of a sequence of admissible task space elements and valid motion path through them.

\begin{algorithm}[h]
\small
	\caption{High-Level Approach}\label{alg:overview}
	\begin{algorithmic}[1]
		\Procedure{Approach}{Task and Motion Planning Problem}
		    \State $\mathcal{S}_\texttt{optimisitic},\mathcal{S}_\texttt{best}\leftarrow\emptyset$
		    \While{\texttt{not converged}}\label{alg:main_loop}
	            \State $\mathcal{H}\leftarrow$\texttt{ExpandRepresentation($\mathcal{H}$)}\label{alg:overview_expand_representation}
                \While{$\mathcal{S}_\texttt{best}$ \texttt{not optimal for} $\mathcal{H}$}\label{alg:overview_nbs_loop}
		            \State $\mathcal{S}_\texttt{optimisitic}\leftarrow$\texttt{ComputeTaskPlan($H$)}\label{alg:overview_compute_task_plan}
		            \State $\mathcal{S}_\texttt{complete}\leftarrow$\texttt{ResolveConflicts($\mathcal{H},\mathcal{S}_\texttt{optimisitic}$)}\label{alg:overview_resolve_conflicts}
		            \If{$\mathcal{S}_\texttt{complete}.\texttt{cost}<\mathcal{S}_\texttt{best}.\texttt{cost}$}\label{alg:overview_check_update_plan}
		                \State$\mathcal{S}_\texttt{best}\leftarrow\mathcal{S}_\texttt{complete}$\label{alg:overview_update_plan}
                    \EndIf
                    \If{\texttt{earlyTermination} and $\mathcal{S}_\texttt{best}\neq\emptyset$}\label{alg:overview_early_termination}
                        \State\Return$S_\texttt{best}$\label{alg:overview_early_return}
		            \EndIf
        		\EndWhile\label{alg:overview_end_nbs}
			\EndWhile
			\State\Return$S_\texttt{best}$\label{alg:overview_return}
		\EndProcedure
	\end{algorithmic}
\end{algorithm}

\textcolor{black}{
We include two variants in Algorithm~\ref{alg:overview}.
An \texttt{earlyTermination} option can be used to return the first found solution (line~\ref{alg:overview_early_return}). 
This does not offer any optimality guarantees though we will show later that it can be probabilistically complete.
}

\textcolor{black}{
For an asymptotically optimal solution, we first continuously expand the representation on line~\ref{alg:overview_expand_representation}. 
Then we use a hybrid approach inspired by the work in~\cite{bpssk-ostaapffmrap-20} to find the optimal solution for the current representation (lines~\ref{alg:overview_nbs_loop}-\ref{alg:overview_end_nbs}) by iteratively computing and validating task plans.
Additionally, anytime behavior can be gained by returning a valid $\mathcal{S}_\texttt{best}$ before convergence.
Details of these variations are discussed in Section~\ref{section:method_variants}.
Probabilistic completeness and asymptotic-optimality are both dependent upon the underlying construction and search methods used in each stage.  
}

\subsection{Task Space Hypergraph}
\label{section:task_space_hypergraph}

\begin{figure}[h]
   \centering
   \includegraphics[width=.8\linewidth]{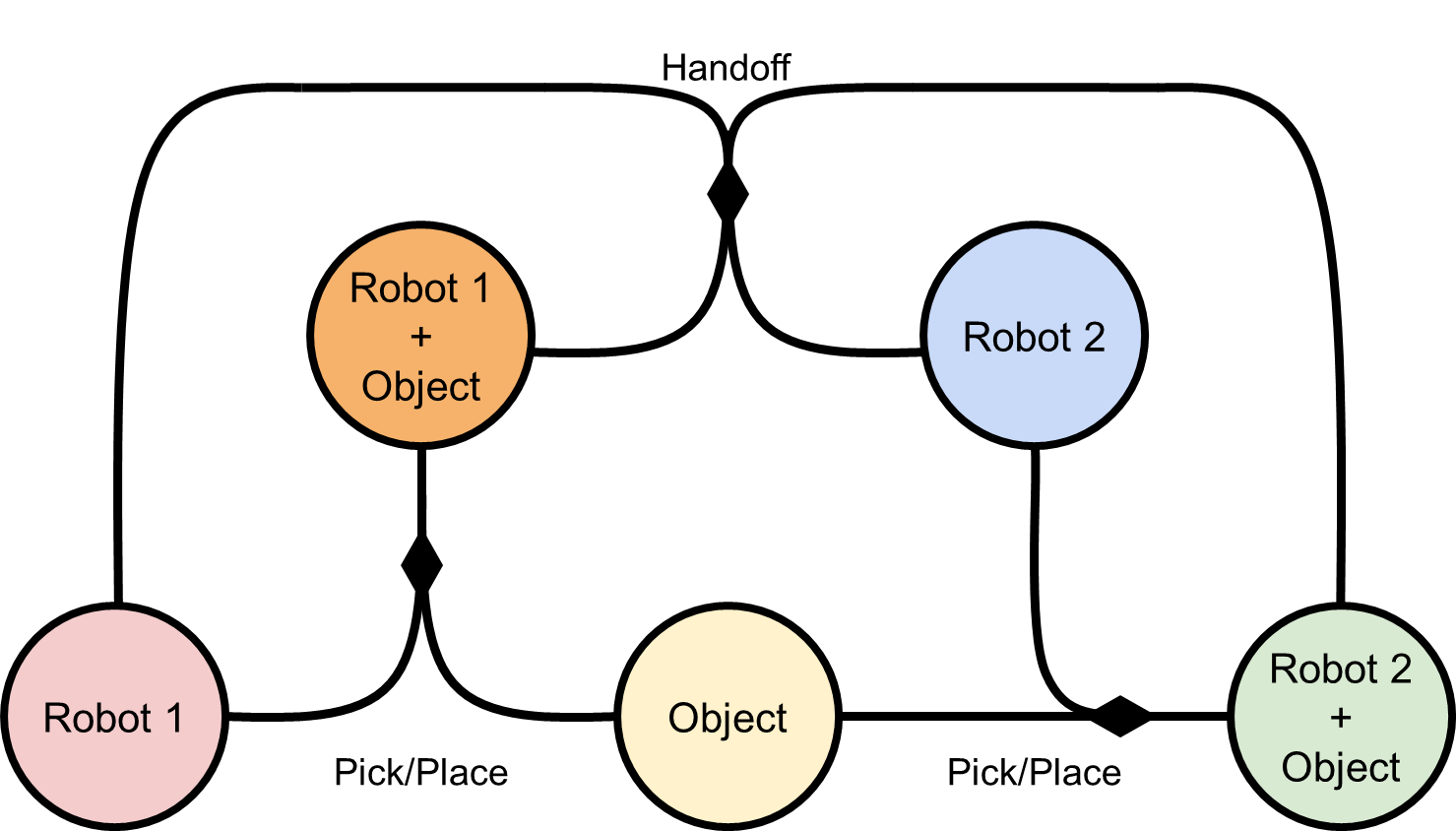} 
   \caption{
   This image depicts the task space hypergraph for a multi-manipulator problem with two robots and one object system.
   The vertices represent the \textcolor{black}{set of task space elements} obtainable in the system.
   The hyperarcs encode the set of \textcolor{black}{allowed transitions} available by applying \textit{pick, place,} and \textit{handoff} actions.
   The set of \textcolor{black}{admissible task space elements} these can combine to create are captured in Fig.~\ref{fig:multi-kg}, and the colors generally correlate.
   The red vertex represents Robot 1 not holding any object.
   The blue vertex represents Robot 2 not holding any object.
   The yellow vertex represents the object at a stable pose.
   The combination of these three vertices creates the \textcolor{black}{admissible task space element} depicted in Fig.~\ref{fig:multi-kg-stable-pose}.
   The hyperarc from the red and yellow vertices to the orange vertex denotes a pick (or place) action and includes only the moveable bodies involved.
   The hyperarc from the orange and blue vertices to the red and green vertices encodes a handoff action (like the one depicted in Fig.~\ref{fig:transition-manip}), and the hyperarc from the green vertex to the yellow and blue vertices encodes a place (or pick) action.
   This sequence is the same as the sequence of \textcolor{black}{admissible task space elements} depicted in Fig.~\ref{fig:multi-kg}, but only the moveable bodies that experience change are involved in the hypergraph transitions. 
   }
   \label{fig:task-space-hypergraph}
\end{figure}
\textcolor{black}{
As defined in Section~\ref{section:prob_def}, task planning can be represented by a \textit{task space} $\mathcal{T}$ where each task space element $T_i=(B_i,\mathcal{W}_i,C_i)\in\mathcal{T}$ includes a set of moveable bodies $B_i\subseteq\mathcal{B}$, the configuration space $\mathcal{W}_i$ for $B_i$, and a set of constraints $C_i$ defining validity in $\mathcal{W}_i$.
Task space elements can be decomposed or combined as defined in Section~\ref{section:prob_def}.
Additionally, transitions can occur between sets of task space elements so long as all moveable bodies present on one side of the transition are present on the other side of the transition.
}

\textcolor{black}{
We propose a \textit{task space hypergraph} $\tsh=(\tsv,\tse)$ to model the set of task space elements and their relationships (Fig.~\ref{fig:task-space-hypergraph}).
Each vertex $v_\mathcal{T}=<T_i>\in\tsv$ encodes a task space element $T_i\in\mathcal{T}$.
Hyperarcs $E_\mathcal{T} = <\texttt{Tail},\texttt{Head}>\in\tse$ encode either a \textit{composition} or \textit{transition} relationship from a tail set of task space elements to a head set of task space elements (Fig.~\ref{fig:transition-manip}).
}

\textcolor{black}{
\textit{Composition hyperarcs}
$E_\mathcal{T}^\texttt{comp}\in\tse$ represent a change in compositions of moveable bodies, their configuration spaces, and constraints.
These may correspond to a decomposition, combination, or a mix of the two.
}

\textcolor{black}{
\textit{Transition hyperarcs} $E_\mathcal{T}^\texttt{trans}$ represent a transition between the configuration spaces and/or constraints for a set of moveable bodies.
For both relationship types, the set of elements in the tail must be independent (i.e. no overlapping moveable bodies).
Similarly, the set of elements in the headset must be independent.
}

\textcolor{black}{
A set of \textit{composition rules} may be used to restrict the allowable decompositions and combinations of task space elements to reduce the size of the task space.
For example, a common choice in MRMP is to only consider task space elements which contain a single robot.
A multi-manipulator planning algorithm may consider only task space elements which contain only robots and objects involved in direct manipulation with each other (e.g. a robot grasping an object).
Transitions then occur between sets of task space elements.
}

\textcolor{black}{
Similarly, constraints on what transitions are feasible between sets of task space elements can restrict arbitrary transitions between configuration spaces or constraint sets for a set of moveable bodies.
These constraints along with the composition rules are packaged as \textit{allowed transitions} $A$, and implicitly define the set of obtainable vertices and hyperarcs within a task space hypergraph.
}

\textcolor{black}{
Alg.~\ref{alg:tsh} defines how the task space hypergraph $\tsh$ can be constructed from the initial task space element $T_\texttt{init}$ and the set of allowed transitions $A$. 
The direction of expansion is implemented in the \texttt{ExpandHypergraph} function and may be configured to a BFS, DFS, or heuristic guided search. An \texttt{earlyQuit} option may be included if a single task solution is sufficient. 
}

\begin{algorithm}
\small
	\caption{Task Space Hypergraph Construction}\label{alg:tsh}
	\begin{algorithmic}[1]
		\Procedure{Build $\tsh$}{$T_\texttt{init}$, Allowed Transitions $A$, \texttt{earlyQuit}}
		    \State$D(T_\texttt{init})\leftarrow\texttt{Decompose}(T_\texttt{init},A)$
		    \State$\tsh=(\tsv,\tse)\leftarrow\texttt{Initialize}(D(T_\texttt{init}),A)$
		    \While{\texttt{true}}
		        \State$\mathcal{V}_\texttt{new},\mathcal{E}_\texttt{new}\leftarrow\texttt{ExpandHypergraph}(\tsh,A)$
		        \If{$\mathcal{E}_\texttt{new}==\emptyset$}
		            \State break
		        \EndIf
		        \State$\tsv\leftarrow\tsv\cup\mathcal{V}_\texttt{new};\tse\leftarrow\tse\cup\mathcal{E}_\texttt{new}$
		        \If{\texttt{earlyQuit} and \texttt{ContainsSolution}($\tsh$)}
		            \State break
		        \EndIf
		    \EndWhile
		    \State\Return$\tsh$
		\EndProcedure
	\end{algorithmic}
\end{algorithm}

\subsection{Motion Hypergraph}
\label{section:motion_hypergraph}

\begin{figure}[h]
   \centering
   \includegraphics[width=.8\linewidth]{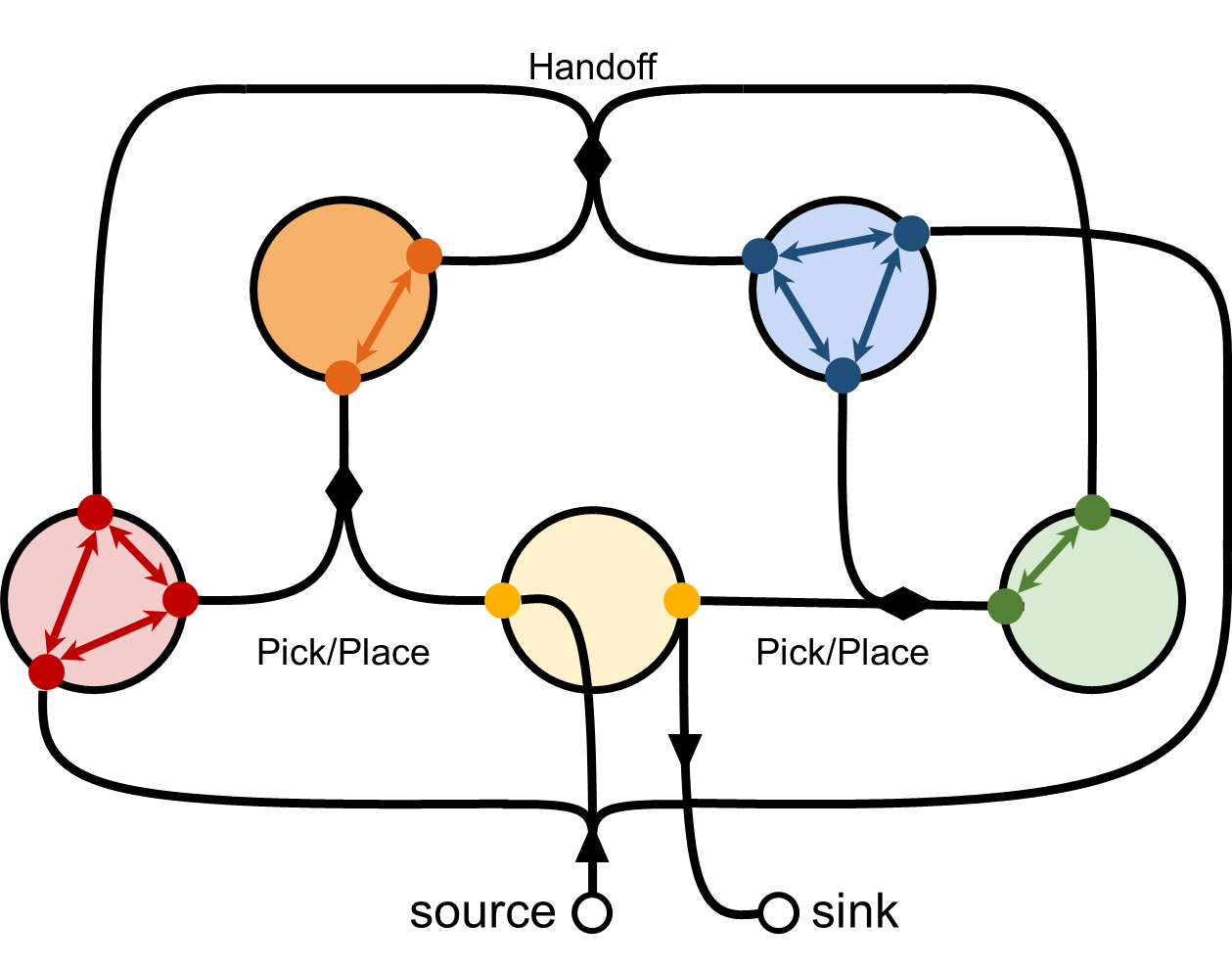} 
   \caption{This motion hypergraph is an augmentation of the multi-manipulator task space hypergraph in Figure~\ref{fig:task-space-hypergraph}.
            \textcolor{black}{Transition hyperarc paths} are planned for each \textcolor{black}{transition hyperarc} in the task space hypergraph.
            The start and end configurations of the \textcolor{black}{transition paths} are represented as colored vertices within the corresponding \textcolor{black}{task space elmements}.
            \textcolor{black}{Move paths} are planned between transition start/end configurations within the same \textcolor{black}{task space element} and represented as (colored) edges to encode movement within that \textcolor{black}{task space element}.
            Note there are no internal edges within the yellow object \textcolor{black}{task space element} as it is unactuated.
   }
   \label{fig:grounded-hypergraph}
\end{figure}

\textcolor{black}{
The task space hypergraph considers only task space conditions and does not consider the motion feasibility of transitions.
We build a \textit{motion hypergraph} $\moh=(\mov,\moe)$ to expand the task space information in the task space hypergraph to include motion feasibility.
This motion hypergraph is a sampled representation of possible transitions through the multi-robot task and motion planning space.
}

\subsubsection{Motion Vertices}
\label{section:motion_vertices}


\textcolor{black}{
The vertices in the task space hypergraph contain task space constraints, but they do not have any {\dof} values to represent an explicit configuration within the configuration space $\mathcal{W}_i$ of the corresponding task space element $T_i$.
A motion vertex $v_\mathcal{M} = <v_{\mathcal{T}},q>\in\mov$ contains both a task space hypergraph vertex $v_{\mathcal{T}}\in\tsv$ (corresponding to a task space element $T_i\in\mathcal{T}$) and an explicit configuration $q\in\mathcal{W}_i$.
As such, each $v_{\mathcal{T}}\in\tsv$ may map to many motion vertices in $\mov$.
}

\subsubsection{Motion Hyperarcs}
\label{section:motion_hyperarcs}

\textcolor{black}{
There are three kinds of hyperarcs in the motion hypergraph: \textit{composition, transition and move.}
}

\textcolor{black}{
A \textit{composition hyperarc} simply indicates that the planning space composition has changed.
}

\textcolor{black}{
A \textit{transition hyperarc} $E_\mathcal{M}^{trans}$ encodes a set of feasible motions for performing a task space transition. 
The tail and head sets of $E_\mathcal{M}^{trans}$ may lie in different task space elements (or combinations of task space elements).
}

\textcolor{black}{
A \textit{move hyperarc} $E_\mathcal{M}^{move}$ encodes a motion path between two motion vertices within the same task space element.
The tail and head sets each contain only a single vertex, $v_\mathcal{M}^\texttt{tail}$ and $v_\mathcal{M}^\texttt{head}$ respectively, and each has the same corresponding task space vertex $v_\mathcal{M}^\texttt{tail}.v_\mathcal{T}$ = $v_\mathcal{M}^\texttt{head}.v_\mathcal{T}$.
They do not correspond to a transition in the task space hypergraph.
}

\textcolor{black}{
Each motion hyperarc $E_\mathcal{M} = <\texttt{Tail},\texttt{Head},E_{\mathcal{T}},\pi>$ consists of the standard tail and head vertex sets, in addition to a corresponding task space hyperarc $E_{\mathcal{T}}\in\tse$, and a path $\pi$ from the motion vertices $v_\mathcal{M}\in E_\mathcal{M}.\texttt{Tail}$ to the motion vertices $v_\mathcal{M}\in E_\mathcal{M}.\texttt{Head}$.
The task space hyperarc $E_{\mathcal{T}}=\texttt{NULL}$ for motion hyperarcs.
The path for composition and transition motion hyperarcs $E_\mathcal{M}^{comp}.\pi$ or $E_\mathcal{M}^{trans}.\pi$ must contain a configuration valid in both the constrained configuration spaces 
produced by combining all task space elements in the tail set $E_\mathcal{M}.\texttt{Tail}$ and the head set $E_\mathcal{M}.\texttt{Head}$.
The path should be valid in the tail set $\cspace$ before the switch and valid in the head set $\cspace$ after.
In the case of composition hyperarcs, the path is often a single configuration.
}

\subsubsection{Composition Hyperarc Planning}
\label{section:composition_hyperarc_plannings}

\textcolor{black}{
A motion composition hyperarc $E_\mathcal{M}^\texttt{comp}\in\moe$ captures the composition change encoded in the task space composition hyperarc $E_\mathcal{T}^\texttt{comp}=E_\mathcal{M}^{comp}.E_\mathcal{T}\in\tse$.
As the constraints do not change (outside the relaxation of collision avoidance between moveable bodies in different task space elements), $E^\texttt{comp}_\mathcal{M}.\pi$ can be any path in the constrained $\cspace$ induced by the combination of the elements in the tail (or head) set of $E_\mathcal{T}^\texttt{comp}\in\tse$.
These paths may be a single configuration which may be sampled, or encode a longer motion.
Paths longer than a single configuration can be planned in the same manner as move hyperarcs operating in the combination of task space elements in the tail (or head) set of $E_\mathcal{T}^\texttt{comp}\in\tse$.
}

\textcolor{black}{
The end points of the path are decomposed along the task space elements present in the tail and head vertices in the task space hyperarc $E_\mathcal{M}^{comp}.E_\mathcal{T}$.
These comprise the tail and head sets of $E_\mathcal{M}^{comp}$ and are included in $\mov$.
$E_\mathcal{M}^{comp}$ is included in $\moe$, encoding a feasible path for performing the composition change.
}

\subsubsection{Transition Hyperarc Planning}
\label{section:transition_hyperarc_plannings}

\textcolor{black}{
Generating transition hyperarcs often requires a specialized planner to capture feasible motions transitioning from the constraints of one set of task space elements to another.
For example, pick, place, and handoff transitions can be computed using an IK based planner.
}

\textcolor{black}{
The $E_\mathcal{M}^{trans}.\pi$ plan can consist of a single configuration or a path.
The start and end points of the path are decomposed 
along the task space elements present in the tail and head vertices in the task space hyperarc $E_\mathcal{M}^{trans}.E_\mathcal{T}$ in the same manner as the compoistion hyperarcs.
The tail and head sets of $E_\mathcal{M}^{trans}$ are included in $\mov$.
$E_\mathcal{M}^{trans}$ is included in $\moe$, encoding a feasible path for performing a task space transition.
}

\subsubsection{Move Hyperarc Planning}
\label{section:move_hyperarc_planning}

\textcolor{black}{
Move hyperarcs encode motion feasibility within task space element configuration spaces $\mathcal{W}_\texttt{move}$ between the start and end points of transition hyperarcs.
While any motion planner may be used to generate the path $E_\mathcal{M}^\texttt{move}.\pi$ between a pair of motion vertices $v_i,v_j$, within the same task space elements $v_i.v_\mathcal{T}=v_j.v_\mathcal{T}$, we chose to utilize a sampling based approach.
Within a particular task space element's configuration space $\mathcal{W}_\texttt{move}$, we grow a roadmap and connect the start and end points of transition hyperarcs within $\mathcal{W}_\texttt{move}$.
Paths between these transition start and end points are found by querying the roadmap.
Each $E_\mathcal{M}^{move}\in\moe$ thus encodes motion feasibility between planned transitions in and out of a particular task substate.
}

\subsubsection{Start and Goal}
\label{section:move_start_and_goal}

\textcolor{black}{
The initial configuration of the system $q\in\mathcal{W}_\texttt{init}$ can be decomposed into a set of motion vertices following the initial task state decomposition $D(T^\texttt{init},A)$ where $A$ is the set of allowed transitions.
These motion start vertices can be connected to transition start points via move hyperarc planning.
Goal constraints can be directly sampled to create motion goal vertices.
These can also be connected to transition end points via move hyperarc planning.
}

\textcolor{black}{
A \textit{virtual source} vertex $v_\mathcal{M}^\texttt{source}$ along with a virtual hyperarc $E_\mathcal{M} = <\{v_\mathcal{M}^\texttt{source}\},\{$all start vertices$\},\texttt{NULL},\emptyset>$ connecting $v_\mathcal{M}^\texttt{source}$ to the set of start vertices is used to capture the entirety of the start state.
If multiple decompositions of the start state are included, then each complete set of start vertices would form the head of a hyperarc originating from the virtual start vertex.
}

\textcolor{black}{
A \textit{virtual target} vertex $v_\mathcal{M}^\texttt{target}$ along with virtual hyperarcs of the form $E_\mathcal{M} = <\{$satisfying set of vertices$\},\{v_\mathcal{M}^\texttt{target}\},\texttt{NULL},\emptyset>$ connecting unique sets of satisfying motion vertices to $v_\mathcal{M}^\texttt{source}$ is used to capture the necessary conditions to complete the task.
}

\subsubsection{Construction}
\label{section:motion_hypergraph_construction}

\textcolor{black}{
The construction of the motion hypergraph $\moh$, defined in Algorithm~\ref{alg:grh}, begins by initializing $\moh$ with the virtual source and target vertices $v_\mathcal{M}^\texttt{source},v_\mathcal{M}^\texttt{target}$ on line~\ref{alg:grh:init}.
The initial set of motion vertices $\mathcal{V}_\texttt{init}$ are computed by decomposing the initial configuration $q^\texttt{init}$ according to the decomposition provided in $D(T_\texttt{init})$ on line~\ref{alg:grh:decomp_start}.
These are added to the hypergraph and connected to $v_\mathcal{M}^\texttt{source}$ on lines~\ref{alg:grh:add_starts},\ref{alg:grh:add_source_hyperarc}.
The same process is repeated for all valid goal configurations provided in $Q^\texttt{goal}$, connecting them to $v_\mathcal{M}^\texttt{target}$ (lines~\ref{alg:grh:decomp_goals1}-\ref{alg:grh:decomp_goals2}).
A \textit{vertex map} $M_v$ tracks all motion instances of task space vertices (lines~\ref{alg:grh:vertex_map1}-\ref{alg:grh:vertex_map2}).
}

\textcolor{black}{
The main loop iterates, first sampling transitions (and compositions) and then moves until the hypergraph contains a solution (line~\ref{alg:grh:loop}).
Transitions are sampled by iterating over all task space hyperarcs $E_\mathcal{T}\in\tse$ (line~\ref{alg:grh:sample_trans1}), and, for some desired number of samples, calling the transition planner to find a feasible path to perform the transition (line~\ref{alg:grh:trans_planner}).
}

\textcolor{black}{
Motion composition hyperarcs may also be sampled here, but due to the number of possible samples, it is often best to lazily discover the need for moving in and out of compositions.
Section~\ref{section:identifying_composition_hyperarcs} discusses this as a means of resolving inter-path conflicts.
}

\textcolor{black}{
The start and end points of successful transitions (or compositions) are added as motion vertices to the hypergraph, and the transition (or compositions) path as a whole is added as a hyperarc (lines~\ref{alg:grh:add_trans_vert},\ref{alg:grh:add_trans_arc}).
The vertex map is updated with the new vertices on line~\ref{alg:grh:update_vertex_map}.
}

\textcolor{black}{
Move hyperarcs are computed by iterating over all of the vertices in the task space hypergraph $v_\mathcal{T}\in\tsv$ and attempting to connect motion vertices from the same task vertex (lines~\ref{alg:grh:sample_move1},\ref{alg:grh:sample_move11}).
A \texttt{Candidates} function on line~\ref{alg:grh:move_cand} allows for a heuristic selection of which vertices to attempt to connect (the default is all other $v_\mathcal{M}\in M_v[v_\mathcal{T}]$).
On line~\ref{alg:grh:move_planner}, the move planner $P_\texttt{move}$ attempts to compute a feasible motion plan between the two motion vertices in the corresponding subconfiguration space.
Successful motion plans are saved as move hyperarcs on line~\ref{alg:grh:add_move_hyperarc}.
}

\textcolor{black}{
The \texttt{singleShot} option may be used if only the task space hyperarcs for a specific solution are provided.
This indicates than any failure to sample a hyperarc will prevent a solution from being contained within $\moh$, and lines~\ref{alg:grh:fail_trans},\ref{alg:grh:fail_move} allow for early termination upon failure.
If all hyperarc samples are successful, then the \texttt{singleShot} options allows for early return of motion hypergraph without mandating that it contain a solution (line~\ref{alg:grh:singleshot_return}).
}

\begin{algorithm}
\small
	\caption{Motion Hypergraph Construction}\label{alg:grh}
	\begin{algorithmic}[1]
		\Procedure{Build $\moh$}{Task Space Hypergraph $\tsh=(\tsv,\tse)$, composition planner $P$, transition planner $P_\texttt{trans}$, move planner $P_\texttt{move}$, initial task state decomposition $D(T_\texttt{init})$, initial configuration $q_\texttt{init}$, goal task state decomposition $D(T_\texttt{goal})$, goal configurations $Q_\texttt{goal}$, \texttt{singleShot}}
		\State Initialize Motion Hypergraph with start and goal configurations
		\State $\moh\leftarrow(\mov,\moe)=(\{v_\mathcal{M}^\texttt{source},v_\mathcal{M}^\texttt{target}\},\emptyset)$\label{alg:grh:init}
		\State$\mathcal{V}_\texttt{init}\leftarrow\texttt{DecomposeCfg}(q^\texttt{init},D(T_\texttt{init}))$\label{alg:grh:decomp_start}
		\State$\mov\leftarrow\mov\cup \mathcal{V}_\texttt{init}$\label{alg:grh:add_starts}
		\State$\moe\leftarrow\moe\cup \{<\{v_\mathcal{M}^\texttt{source}\},\mathcal{V}_\texttt{init},\texttt{NULL},\emptyset>$\}\label{alg:grh:add_source_hyperarc}
		\ForAll{$q^\texttt{goal}\in Q^\texttt{goal}$}\label{alg:grh:decomp_goals1}
		    \State$\mathcal{V}_\texttt{goal}\leftarrow\texttt{DecomposeCfg}(q^\texttt{goal},D(T_\texttt{goal}))$
    		\State$\mov\leftarrow\mov\cup \mathcal{V}_\texttt{goal}$
    		\State$\moe\leftarrow\moe\cup \{\mathcal{V}_\texttt{goal},<\{v_\mathcal{M}^\texttt{target}\},\texttt{NULL},\emptyset>$\}
		\EndFor\label{alg:grh:decomp_goals2}
		\State vertex map $M_v\leftarrow\emptyset$\label{alg:grh:vertex_map1}
		\ForAll{$v_\mathcal{M}\in\mov$}
		    \State $M_v[v_\mathcal{M}.v_\mathcal{T}]\leftarrow M_v[v_\mathcal{M}.v_\mathcal{T}]\cup \{v_\mathcal{M}\}$
		\EndFor\label{alg:grh:vertex_map2}
		\While{does not \texttt{ContainsSolution}($\moh$)}\label{alg:grh:loop}
    		\State Sample Transitions (and Compositions)
    		\ForAll{$E_\mathcal{T}\in\tse$}\ForAll{desired number of samples}\label{alg:grh:sample_trans1}
    		    \State $E_\mathcal{M}\leftarrow P_\texttt{trans}(E_\mathcal{T})$ (or $P_\texttt{comp}(E_\mathcal{T})$\label{alg:grh:trans_planner})
    		    \If{$E_\mathcal{M}==\emptyset$}
	                \If{\texttt{singleShot}}\label{alg:grh:fail_trans}
	                    \State\Return$\moh$
	                \Else\State continue
	                \EndIf
	            \EndIf
    		    \State$\mov\leftarrow\mov\cup E_\mathcal{M}.\texttt{Tail}\cup E_\mathcal{M}.\texttt{Head}$\label{alg:grh:add_trans_vert}
    		    \State$\moe\leftarrow\moe\cup \{E_\mathcal{M}\}$\label{alg:grh:add_trans_arc}
    		    \ForAll{$v_\mathcal{M}\in E_\mathcal{M}.\texttt{Tail}\cup E_\mathcal{M}.\texttt{Head}$}
    		        \State $M_v[v_\mathcal{M}.v_\mathcal{T}]\leftarrow M_v[v_\mathcal{M}.v_\mathcal{T}]\cup \{v_\mathcal{M}\}$\label{alg:grh:update_vertex_map}
    		    \EndFor
    		\EndFor\EndFor\label{alg:grh:sample_trans2}
    		\State Sample Moves
    		\ForAll{$v_\mathcal{T}\in\tsv$}\label{alg:grh:sample_move1}
    		    \ForAll{$v_{g,i}\in M_v[v_\mathcal{T}]$}\label{alg:grh:sample_move11}
    		        \ForAll{$v_{g,j}\in\texttt{Candidates}(v_{g,i},M_v[v_\mathcal{T}])$}\label{alg:grh:move_cand}
    		            \State$E_\mathcal{M}\leftarrow P_\texttt{move}(v_{g,i},v_{g,j})$\label{alg:grh:move_planner}
    		            \If{$E_\mathcal{M}==\emptyset$}
    		                \If{\texttt{singleShot}}\label{alg:grh:fail_move}
    		                    \State\Return$\moh$
    		                \Else\State continue
    		                \EndIf
    		            \EndIf
    		            \State$\moe\leftarrow\moe\cup\{E_\mathcal{M}\}$\label{alg:grh:add_move_hyperarc}
    		        \EndFor
    		    \EndFor\label{alg:grh:sample_move2}
    		\EndFor
    		\If{\texttt{singleShot}}\label{alg:grh:singleshot_return}
    		    \State\Return$\moh$\label{alg:grh:returnearly}
    		\EndIf
    	\EndWhile
    	\State\Return$\moh$\label{alg:grh:return}
		\EndProcedure
	\end{algorithmic}
\end{algorithm}

\subsection{Transition-extended Hypergraph}
\label{section:transition_extended_hypergraph}

\begin{figure}[h]
\centering
 \subfloat[Valid History]{
	\includegraphics[width=.45\linewidth]{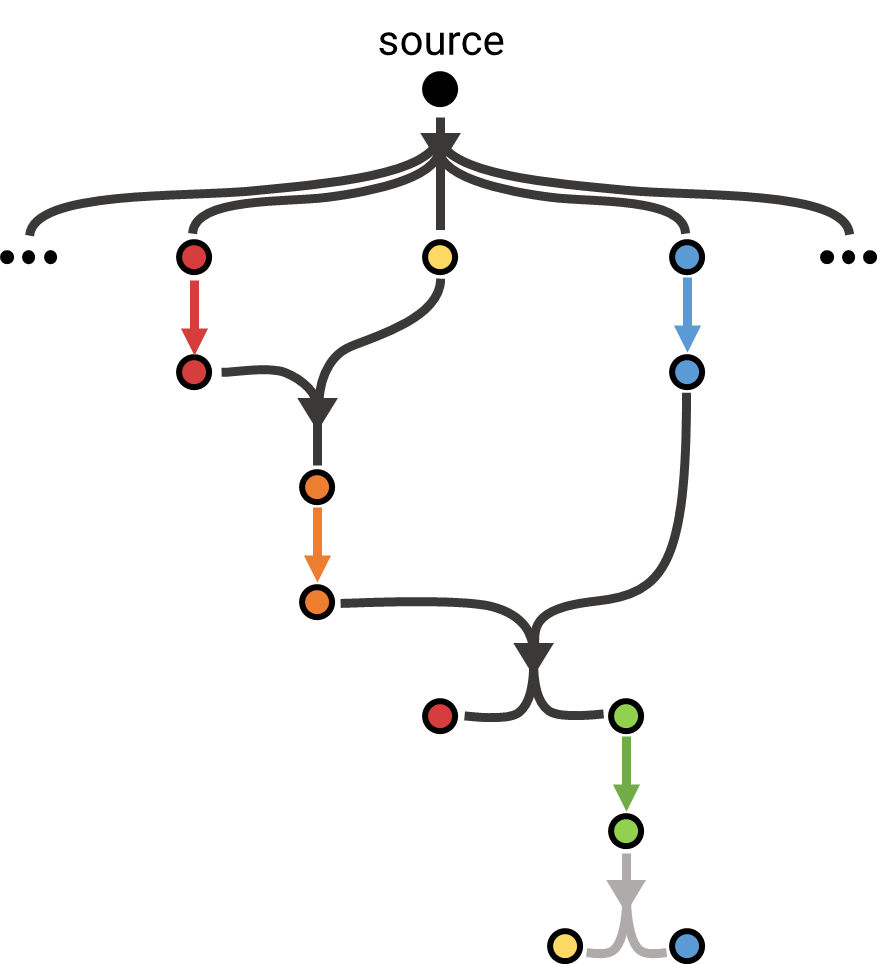}
	\label{fig:manip-transition-extended-hypergraph-valid}
}
 \subfloat[Invalid History]{
	\includegraphics[width=.45\linewidth]{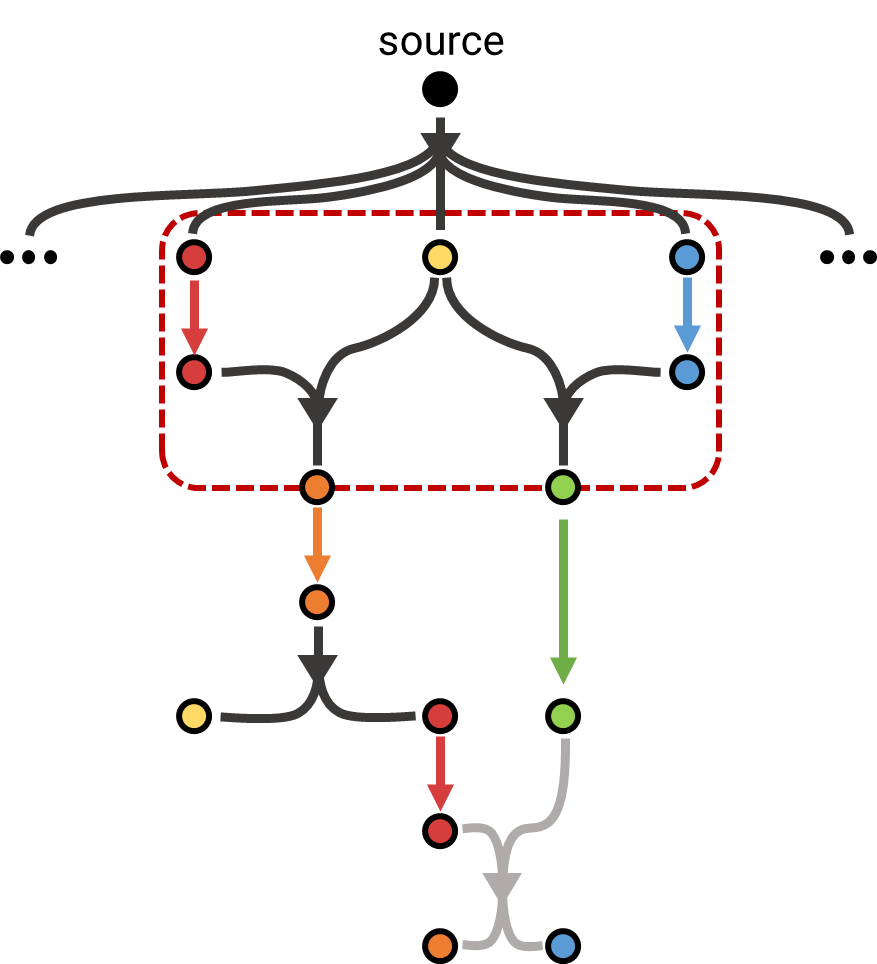}
	\label{fig:manip-transition-extended-hypergraph-invalid}
}
\caption{
	Both figures (a,b) consider a transition-extended version of the multi-manipulator space depicted in the task space hypergraph (Fig.~\ref{fig:task-space-hypergraph}) and motion hypergraph (Fig.~\ref{fig:grounded-hypergraph}) where robot 2 can now reach the object's starting position and there may be more robots and objects involved outside the vertices and hyperarcs depicted.
	The colors of the vertices correspond to the previous hypergraphs (yellow-object, red-robot 1, blue-robot 2).
    In either figure, we consider the transition history for the tail set of the light gray hyperarc.
    Figure (a) depicts a valid transition history where no vertex has more than one outgoing hyperarc.
    Figure (b) depicts an invalid transition history.
    The conflict in the transition history is in the dotted red box.
    By including the pair of outgoing hyperarcs of the yellow object vertex in the same transition history, the object must exist in two conflicting states at once.
}
\label{fig:manip-transition-extended-hypergraphs}
\end{figure}

\textcolor{black}{Because both transition and composition hyperarcs in the motion hypergraph $\moh$ move in and out of different space compositions, planning directly over $\moh$ can result in a moveable body existing in multiple states simultaneously.}
In order to query a valid task plan, we must consider a \textit{transition-extended} search space which we represent with a \textit{transition-extended hypergraph} $\teh=(\tev,\tee)$.

\subsubsection{Transition-extended Vertices}
\label{section:transition_extended_vertices}

Each vertex $v_\texttt{TE}=<v_\mathcal{M},\Pi_{v_\texttt{TE}^\texttt{source},v_\texttt{TE}}>\in\tev$ is defined by a motion vertex $v_\mathcal{M}\in\mov$ and a \textit{transition history} $\Pi_{v_\texttt{TE}^\texttt{source},v_\texttt{TE}}$.

$\teh$ has a virtual source vertex $v_\texttt{TE}^\texttt{source}=<v_\mathcal{M}^\texttt{source},\Pi_{v_\texttt{TE}}^\texttt{source},v_\texttt{TE}^\texttt{source}>$ which contains a trivial transition history.
The transition history of every other vertex $v_\texttt{TE}\in\tev/\{v_\texttt{TE}^\texttt{source}\}$ consists of all of the moves and transitions taken to reach $v_\texttt{TE}$ from $v_\texttt{TE}^\texttt{source}$.
This corresponds to a hyperpath $\Pi_{v_\texttt{TE}^\texttt{source},v_\texttt{TE}}$ in $\teh$ from $v_\texttt{TE}^\texttt{source}$ to $v_\texttt{TE}$.
As such, each vertex $v_\texttt{TE}$ has a single incoming hyperarc as this incoming hyperarc is part of its transition history.

A transition history $\Pi_{v_\texttt{TE}^\texttt{source},v_\texttt{TE}}$ is valid if every vertex $v_\texttt{TE}\in\Pi_{v_\texttt{TE}^\texttt{source},v_\texttt{TE}}$ contains at most one outgoing hyperarc.
This indicates that the moveable bodies at a particular configuration denoted by the vertices in the tail set of a hyperarc transition to at most one additional configuration at a time.
Multiple outgoing hyperarcs from a single transition-extended vertex $v_\texttt{TE}$ indicates that the moveable bodies corresponding to $v_\texttt{TE}$ perform simultaneous motions and exist in more than one place at once.
This is an invalid transition history.

In addition to ensuring that each moveable body performs at most one motion at once, the transition history also encodes the time each transition or move is expected to occur and the ``last known position'' for each moveable body according to the set of transitions and moves in the transition history.

\subsubsection{Transition-extended Hyperarcs}
\label{section:transition_extended_hyperarcs}

Each hyperarc $E_\texttt{TE}=<\texttt{Tail},\texttt{Head},E_\mathcal{M}>$ corresponds to either a composition, transition or move hyperarc $E_\mathcal{M}\in\moe$.
The head set consists of vertices which map to the motion vertices in the motion head set of $E_\mathcal{M}$.
The tail set consists of the union of vertices which map to the motion tail set of $E_\mathcal{M}$ and a set of \textit{scheduling constraint} vertices.

A set of transition-extended vertices 
can form a tail set if the union of their transition histories is valid.
This is necessary for the head set vertices of $E_\texttt{TE}$ to have valid transition histories.

\subsubsection{Goal Vertices}
\label{section:transition_extended_hypergraph_goals}

Any transition extended vertex $v_\texttt{TE}$ such that $v_\texttt{TE}.v_\mathcal{M}==v_\mathcal{M}^\texttt{target}$ is considered a valid goal vertex.
As each $v_\texttt{TE}\in\tev$ contains a valid and unique transition history, the transition history of goal vertex corresponds to a unique \textit{optimistic schedule} of compositions, transitions, and moves to complete the task.
This schedule is optimistic as while motions are valid within their own subconfiguration spaces, they are not guaranteed to be collision-free with respect to the paths of other moveable bodies occurring simultaneously.
This requires the use of the conflict resolution stage (discussed in Section~\ref{section:conflict-resolution}) which seeks to adjust individual motion plans to resolve conflicts.

\subsubsection{Scheduling Constraints}
\label{section:scheduling_constraints}


\textcolor{black}{
Some motion conflicts will not be resolvable by the conflict resolution layer.
This occurs when the path or position of one moveable body cannot occur simultaneously with the path or position of another moveable body.
These are motion-based \textit{scheduling conflicts}.
}

\textcolor{black}{
Scheduling conflicts can occur either between a pair of motion vertices, a pair of motion hyperarcs, or a vertex and a hyperarc.
\textit{Vertex-vertex conflicts} indicate that once one vertex has been reached, the other cannot be reached until the first one has been left.
This occurs when the configurations for the two motion vertices are in collision.
}

\textcolor{black}{
\textit{Hyperarc-hyperarc conflicts} indicate that the corresponding paths cannot occur simultaneously, and one must be concluded before the other can begin.
This occurs when the motions within separate subconfiguration spaces are found to be incompatible with each other.
}

\textcolor{black}{
\textit{Vertex-hyperarc conflicts} indicate the configuration of a motion vertex completely impedes the path of the hyperarc.
For example, this may occur if a stationary object is blocking the motion of robot.
}

\textcolor{black}{
The conflict resolution stage (Section~\ref{section:conflict-resolution}) may discover these scheduling conflicts and pass the information back to the representation layers as a \textit{scheduling constraint} associated with a motion vertex or hyperarc.
Scheduling constraints indicate that the associated motion vertex/hyperarc pair cannot be included in the hyperpath simultaneously.
Transition histories must satisfy the scheduling constraints of the included motion vertices/hyperarcs to be valid.
Scheduling vertices can be included in a tail set for a transition-extended hyperarc $E_\texttt{TE}$ to augment the transition history with additional transitions and motions which satisfy the scheduling constraint.
In the case of an object blocking the path of a robot, this encodes that the object must not be at the blocking location during the execution of that path of the robot.
}

\subsubsection{Identifying Composition Hyperarcs}
\label{section:identifying_composition_hyperarcs}

\textcolor{black}{
Outside of introducing scheduling constraints, some motion conflicts can be addressed by adding a composition hyperarc to the task space hypergraph $\tsh$ which provides access to the composite space of the conflicting sets of moveable bodies.
This requires the update of the motion and transition-extended hypergraph layers as well to fully encode the composite shift and allow the conflict to be resolved in the composite space.
This resembles the approach taken by M*~\cite{wc-sefmpp-15} upon discovering conflicts.
}

\subsubsection{Implicit Construction and Search}
\label{section:transition_extended_hypergraph_construction}

As the purpose of this graph is to find an optimistic schedule in the form of a goal vertex, and $\teh$ is potentially infinitely large,
we search an implicit representation of $\teh$ and construct the necessary vertices and hyperarcs as part of the search process.
This search, described in Algorithm~\ref{alg:teh}, is an adaptation of the optimal shortest hyperpath query algorithm presented in ~\cite{bpssk-ostaapffmrap-20} to account for the implicit representation and leverage the particular constraints of the transition-extended hypergraph (e.g. each vertex has only a single incoming hyperarc).

The search starts by initializing the source vertex $v_\texttt{TE}^\texttt{source}$ and the explicit $\teh$ on lines~\ref{alg:teh:source},\ref{alg:teh:init}
along with the vertex weight function $W$, the frontier queue $Q$, and the parent map from vertices to incoming hyperarcs $P_v$ on lines~\ref{alg:teh:weight},\ref{alg:teh:queue_parent}.

A set of \textit{partial hyperarcs} $U$ is initialized on line~\ref{alg:teh:partials}.
Due to the implicit (and potentially infinite) nature of the hypergraph, the \textit{forward star} (\texttt{FS}), or set of all outgoing hyperarcs, of a transition-extended vertex $v_\texttt{te}$ cannot always be computed.
This occurs when the forward star of the corresponding motion vertex $\texttt{FS}(v_\texttt{te}.v_\mathcal{M})$ includes hyperarcs with tail sets containing more than one vertex
as any set of transition-extensions of the motion tail set with non-conflicting transition histories may form a tail set for a transition-extended hyperarc.
Instead, we track partial hyperarcs in $U$, where we dynamically construct valid tail sets as new transition-extended vertices are discovered.

When the search attempts to expand from a particular vertex $u$ on line~\ref{alg:teh:fs}, the set of all completed hyperarcs in the forward start of $u$ are expanded.
This maintains the search properties in the original hyperpath query algorithm in~\cite{glpn-dhaa-93} which only expands a hyperarc once all of the vertices in its tail set have been expanded.
The set of partial (and completed) hyperarcs $U$ is updated on line~\ref{alg:teh:partials_update} whenever a new vertex is added to the frontier queue.

For each complete hyperarc $E_\texttt{TE}$ in the forward star of $y$, the search adds $E_\texttt{TE}$ to the hypergraph on line~\ref{alg:teh:add_hyperarc} and computes the maximum weight of the vertices in the tail set on line~\ref{alg:teh:path_weight}.
In the context of planning, this naturally captures any waiting time required by the moveable bodies in one tail vertex on the moveable bodies in the other tail vertices to arrive before beginning the transition.

Next, each of the vertices in the head set of $E_\texttt{TE}$ are added to the hypergraph and the frontier queue (lines~\ref{alg:teh:add_vertex},\ref{alg:teh:push}) before setting the weight and parentage of each vertex (lines~\ref{alg:teh:weight_update},\ref{alg:teh:parent_update}).
Finally, a check if performed on line~\ref{alg:teh:check_goal} to see if any of the head set vertices correspond to the motion target vertex $v_\mathcal{M}^\texttt{target}$, indicating that an optimistic schedule as has been found.
If the queue is empty, and no schedule has been found, a \texttt{NULL} solution is returned.

\begin{algorithm}
\small
	\caption{Transition-extended Hypergraph Search}\label{alg:teh}
	\begin{algorithmic}[1]
		\Procedure{Search $\teh$}{Motion Hypergraph $\moh$, $v_\mathcal{M}^\texttt{source}$, $v_\mathcal{M}^\texttt{target}$}
		    \State$v_\texttt{TE}^\texttt{source}\leftarrow<v_\mathcal{M}^\texttt{source},\Pi_{v_\texttt{TE}^\texttt{source},v_\texttt{TE}^\texttt{source}}>$\label{alg:teh:source}
		    \State$\teh\leftarrow(\{v_\texttt{TE}^\texttt{source}\},\emptyset)$\label{alg:teh:init}
		    \State Weight function $W(v_\texttt{TE}^\texttt{source})\leftarrow0$\label{alg:teh:weight}
		    \State Queue $Q\leftarrow\{v_\texttt{TE}^\texttt{source}\}$; Parent map $P_v\leftarrow\emptyset$\label{alg:teh:queue_parent}
		    \State\textcolor{gray}{Partial hyperarcs $U\leftarrow\emptyset$}\label{alg:teh:partials}
		    \While{$Q\neq\emptyset$}\label{alg:teh:loop}
		        \State$u\leftarrow Q.\texttt{pop}()$\label{alg:teh:pop}
		        \ForAll{$E_\texttt{TE}\in\texttt{FS}_{\textcolor{gray}{complete}} (u,\textcolor{gray}{U})$}\label{alg:teh:fs}
		            \State$\tee\leftarrow\tee\cup\{E_\texttt{TE}\}$\label{alg:teh:add_hyperarc}
		            \State$f\leftarrow\texttt{max}_{v_\texttt{TE}\in E_\texttt{TE}.\texttt{Tail}} W(v_\texttt{TE})$\label{alg:teh:path_weight}
		            \ForAll{$y\in E_\texttt{TE}.\texttt{Head}$}\label{alg:teh:vertices}
		                \State $\tev\leftarrow\tev\cup\{y\}$\label{alg:teh:add_vertex}
		                \State $Q\leftarrow Q\cup\{y\}$\label{alg:teh:push}
		                \State $W(y)\leftarrow f + w(E_\texttt{TE})$\Comment{$w(E_\texttt{TE})$ is the weight} of $E_\texttt{TE}$ \label{alg:teh:weight_update}
		                \State $P_v[y] = E_\texttt{TE}$\label{alg:teh:parent_update}
		                \If{$y.v_\mathcal{M} == v_\mathcal{M}^\texttt{target}$}\label{alg:teh:check_goal}
		                    \State\Return $y$\label{alg:teh:return_solution}
		                \EndIf
		                \State $U\leftarrow\texttt{UpdatePartialHyperarcs}(y)$\label{alg:teh:partials_update}
		            \EndFor
		        \EndFor
		    \EndWhile
		    \State\Return\texttt{NULL}
		\EndProcedure
	\end{algorithmic}
\end{algorithm}

\subsection{Conflict Resolution}
\label{section:conflict-resolution}
The optimistic scheduled produced by searching $\teh$ corresponds to scheduled motions which are only guaranteed to be feasible within their own configuration spaces.
Thus, there is no guarantee that these motions are conflict free with each other.
We address this with a conflict resolution layer.

\textcolor{black}{
Each motion within the schedule is defined by a start and a goal for the corresponding subset of the moveable bodies along with any precedence constraints defined by the schedule.
This resembles a scheduled variant of the multi-robot motion planning problem discussed in Section~\ref{section:related_work_mrmp}.
As such, scheduled variations of multi-robot motion planning methods may be adapted to perform conflict resolution with different theoretical properties.
}

\textcolor{black}{
For fast, single shot planning, priority-based decoupled methods such as Priority-Based Search~\cite{mhslk-swcpfmpf-19} can provide quick solutions.
When an optimal search over the current representation is needed, hybrid methods such as CBS-MP~\cite{smsa-rmmpucs-21} can be adapted to account for scheduling constraints in a manner similar to the grid-world scheduled CBS variant in~\cite{bpssk-ostaapffmrap-20}.
}

\subsection{Method Variants}
\label{section:method_variants}

\textcolor{black}{
The various components of the DaSH method can be configured to balance a trade off in planning times vs solution quality.
On the two extremes, we have the option to return the first valid solution found or to continue searching until an asymptotically optimal solution is found.
}

\subsubsection{Faster Planning Times vs Higher Solution Quality}

\textcolor{black}{
The high-level algorithm described in Alg.~\ref{alg:overview} can be configured to return the first valid solution by setting \texttt{earlyTermination=True} from the beginning.
Alternatively, permanently leaving \texttt{earlyTermination=False} will allows the algorithm to converge on the optimal solution, so long as the \texttt{ExpandRepresentation} function maintains asymptotically optimal properties and the loop in lines~\ref{alg:main_loop}-\ref{alg:overview_end_nbs} returns the optimal solution for the current representation.
The algorithm can be configured with anytime properties if \texttt{earlyTermination} is instead a function which signals that time is up and the current best solution is returned.
}

\subsubsection{Representation Construction}

\textcolor{black}{
As the representation construction is a hierarchical construction process, the design choices at each level can impact the properties of the algorithm as a whole.
The task space hypergraph $\tsh$ construction can be modified by changing the behavior of the functions in Alg.~\ref{alg:tsh}.
If \texttt{earlyQuit=True}, then choosing a greedy \texttt{ExpandHypergraph} function can lead to a small, concise $\tsh$ containing the minimum number of task space elements and transitions to move from the start element to the goal element.
This can be beneficial when the task space is very expansive, and there is not a set of allowable transitions which can keep the size down.
Alternatively, if the problem does have an effective set of allowable transitions, then a BFS style \texttt{ExpandHypergraph} function and \texttt{earlyQuit=False} can quickly build a complete $\tsh$.
So long as \texttt{ExpandHypergraph} is configured to continue expanding $\tsh$ everytime is is called until $\tsh$ contains a complete representation of $\mathcal{T}$ (i.e. all admissible task space elements can be formed by combining elements represented in $\tsh$), probabilistic completeness, asymptotic optimality, and anytime properties are still obtainable. 
}

\textcolor{black}{
The size of the $\tsh$ directly affects the effort of Alg.~\ref{alg:grh} as each hyperarc in the $\tsh$ may be sampled many times (lines~\ref{alg:grh:trans_planner},\ref{alg:grh:move_planner}).
The theoretical properties of the motion hypergraph $\moh$ are determined by the $\tsh$ Alg.~\ref{alg:grh} given and the transition, composition, and move planners.
If the move planner is probabilistically complete and asymptotically optimal within each of the task space elements (e.g. PRM*~\cite{kf-isbaomp-10}), and the transition and composition planners continue to sample new hyperarcs, the motion hypergraph will converge to the optimal set of motions for the given $\tsh$.
}

\subsubsection{Next-Best Search}
\label{section:nbs}

\textcolor{black}{
Asymptotic optimally requires an optimal query of the current representation in lines\ref{alg:overview_nbs_loop}-\ref{alg:overview_end_nbs} of Alg~\ref{alg:overview}.
We use the Next-Best Search algorithm (Alg.~\ref{alg:nbs}) to iteratively build and validate task plans until it converges to the optimal solution for the current representation.
}

\textcolor{black}{
Each task plan represents an \textit{optimistic schedule} of independent motions not guaranteed to be collision-free of each other.
The cost of this optimistic schedule serves as a lower-bound for the solution cost as the solution cost can only increase from resolving conflicts.
The algorithm requires each successive call to \texttt{FindTaskPlan} (line \ref{alg:nbs:find_task_plan}) to return the \textit{next-best} task plan in terms of cost (i.e. each optimistic schedule cost must be at least as much as the previous schedule cost).
This allows each successive iteration to update the lower bound of the solution.
}

\textcolor{black}{
Each optimistic schedule must be converted into a valid solution by resolving any potential conflicts.
The \texttt{ConflictResolution} function (line \ref{alg:nbs:conflict_resolution}) uses the roadmaps in the motion hypergraph along with the schedule to construct a set of collision free paths following the scheduled generated by \texttt{FindTaskPlan}.
If the cost of this valid solution is lower than the previous upper bound, we update the upper bound to reflect this solution and save this as the best solution so far (lines \ref{alg:nbs:if_new_upperbound},\ref{alg:nbs:update_upperbound}).
If no valid solution can be found in the current representation, the cost is treated as infinite, and the upper bound is unchanged.
Schedules are saved in $\mathcal{S}$ (line~\ref{alg:nbs:save_solution})and given to the task planner to ensure that it returns the next-best solution (line~\ref{alg:nbs:find_task_plan}).
}

\textcolor{black}{
The algorithm iterates until an optimistic schedule is produced with a cost greater than the current upper bound (line \ref{alg:nbs:main_loop}).
At this point, the current best solution is determined to be the optimal solution with respect to the current representation, and this solution is returned (line \ref{alg:nbs:return}).
}

This iteration follows the strategy proposed by~\cite{bpssk-ostaapffmrap-20}.
If it fails to discover a valid plan, it returns to the representation construction and expands the representation.

\begin{algorithm}
\small
	\caption{Next-Best Search}\label{alg:nbs}
	\begin{algorithmic}[1]
		\Procedure{NBS}{Motion Hypergraph $\moh$}
			\State$(S^*,T^*)\leftarrow(\emptyset,\infty)$\Comment{Best solution}\label{alg:nbs:init_solution}
			\State$\underline{T}\leftarrow0$\Comment{best lower-bound}\label{alg:nbs:init_lowerbound}
			\State$\mathcal{S}\leftarrow\emptyset$\Comment{Set of discovered solutions}\label{alg:nbs:discovered,solutions}
			\While{$T^*>\underline{T}$}\label{alg:nbs:main_loop}
				\State $(P,\underline{T})\leftarrow$\texttt{FindTaskPlan($\moh,\mathcal{S}$)}\label{alg:nbs:find_task_plan}
				\If{$T^*>\underline{T}$}\label{alg:nbs:if_update_lowerbound}
					\State$(S,T)\leftarrow$\texttt{ConflictResolution($\moh,P$)}\label{alg:nbs:conflict_resolution}
					\If{$T<T^*$}\label{alg:nbs:if_new_upperbound}
						\State$(S^*,T^*)\leftarrow(S,T)$\label{alg:nbs:update_upperbound}
					\EndIf
				\EndIf
				\State$\mathcal{S}\leftarrow\mathcal{S}\cup\{P\}$\label{alg:nbs:save_solution}
			\EndWhile
			\State\Return$S*$\label{alg:nbs:return}
		\EndProcedure
	\end{algorithmic}
\end{algorithm}

\subsubsection{Task Planning}

\textcolor{black}{
In order for Alg.~\ref{alg:nbs} to return the optimal solution for the current representation, the \texttt{FindTaskPlan} function must return the \textit{next-best} solution every time it is called.
This requires it to return the optimal solution the first time it is called, and each solution cost serves as a lower bound for the next call.
Fortunately, the transition-extended hypergraph $\teh$ makes this simple as each vertex contains the unique transition history to itself, a satisfying goal vertex represents a unique solution.
If Alg.~\ref{alg:teh} finds an optimal solution, then the \textit{next-best} solution can be obtained by removing the previous solution from $\teh$ and calling the search again.
This also reduces the cost of computing subsequent solutions if the search frontier is maintained.
We will discuss two optimal variants of Alg.~\ref{alg:teh} and a greedy alternative.
}

\textcolor{black}{
\textbf{Dijkstra-like Hyperpath Query:}
The first option resembles a Dijkstra shortest path search over $\teh$ and follows Alg.~\ref{alg:teh} explicitly.
The frontier queue $Q$ is ordered on the weight $W(y)$ of vertices (line~\ref{alg:teh:weight_update}).
}

\textcolor{black}{
\textbf{A*-like Hyperpath Query:}
\label{section:a-star}
An A*-like search can be configured by adding a lower bound cost-to-go heuristic value to each vertex $v_\texttt{TE}$ in the frontier queue $Q$, and ordering the queue by the sum of the weight and cost-to-go heuristic.
We consider a cost-to-go heuristic derived from path (not hyperpath) distances as described in Section~\ref{section:directed-hyperpath} on the motion hypergraph from $v_\texttt{TE}.v_\mathcal{M}$ to $v^\texttt{sink}_\mathcal{M}$.
This path must only consist of vertices for which the corresponding moveable bodies $v_\texttt{TE}.v_\mathcal{M}.v_\mathcal{T}.T_i.B_i$ are specified in the goal constraints of the problem.
The heuristic value of vertices without goal specified moveable bodies is 0, and they cannot be included in the heuristic path.
In object rearrangement problems, this is equivalent to computing the path distance for an object to its goal state from a given motion hypergraph vertex while ignoring any robot only vertices and hyperarcs.
This provides a lower bound by assuming that all other dependencies happen in parallel before this goal constraint is satisfied (e.g. all robots are immediately available to perform actions and all other objects have already reached their goal positions).
}

\textcolor{black}{
\textbf{Greedy Hyperpath Query:}
\label{section:greedy-search}
If the optimal solution is not required, and a solution is desired as fast as possible, then a greedy search can be used in place of Alg.~\ref{alg:teh}.
The simplest greedy option iteratively constructs a valid transition history one hyperarc at a time.
This approach benefits from a heuristic for choosing the next hyperarc to add, and it may be forced to backtrack if the choices made do not allow for a feasible solution to be reached.
So long as this search is complete, then probabilistic completeness may be maintained if it is supported by the other choices.
}

\subsubsection{Conflict Resolution}

\textcolor{black}{
The variants of conflict resolution are discussed in Section~\ref{section:conflict-resolution}.
An optimal option such as the scheduled variant of CBS-MP~\cite{smsa-rmmpucs-21} is required for Alg.~\ref{alg:nbs} to return an optimal solution.
Otherwise, the fastest option may be used though the completeness affects the probabilistic completeness property of the configuration.
}

\textcolor{black}{
Section~\ref{section:experiments} evaluates an asymptotically optimal configuration using NBS with A* variant of Alg.~\ref{alg:teh} and fast, greedy approach which uses the Greedy Hypergraph Query to generate a single task plan for which it attempts to resolve conflicts before expanding the representation.
Both methods use the scheduled variant of CBS-MP~\cite{smsa-rmmpucs-21} to resolve conflicts.
}

\section{Application to Multi-Robot Motion Planning}
\label{section:mrmp}
\textcolor{black}{
We examine the application of the DaSH method to the multi-robot motion planning problem (MRMP-DaSH) to illustrate the concepts discussed in the previous sections and show how the proposed framework generalizes many existing MRMP methods.
We first formally define the problem: Given a set of robots $\mathcal{B}$ and an environment, find continuous collision-free paths for each robot $r_i\in\mathcal{B}$ from a given start $s_i$ to a given goal $g_i$ in the environment.
}

\textcolor{black}{
The task space $\mathcal{T}_\texttt{MRMP}$ consists of elements $T_i=(B_i,\mathcal{W}_i,C_i)$ where $B_i\subseteq\mathcal{B}$, $\mathcal{W}_i$ is the configuration space for $B_i$, and $C_i$ constrains each $b_j\in B_i$ to not be in collision with any obstacle or any $b_k\in B_k, b_k\neq b_j$.
There is a single admissible task space element $\mathcal{T}^*=\{T_\texttt{complete}=(B_\texttt{COMPLETE},\mathcal{W}_\texttt{COMPLETE},C_\texttt{COMPLETE})\}$ where $B_\texttt{COMPLETE}=\mathcal{B}$ and $C_\texttt{COMPLETE}$ contains all collision avoidance constraints.
}

\textcolor{black}{
The hypergraph representation layers defined in Sections~\ref{section:task_space_hypergraph}-~\ref{section:transition_extended_hypergraph} can be configured by adjusting the allowed transitions $A$ to create the different planning spaces commonly considered by MRMP methods. 
}

\subsection{Composite Representation}
\textcolor{black}{
MRMP-DaSH can be reduced to composite MRMP methods such as ~\cite{sl-uppccdpmrs-2002,ssh-faniaehdrfeoirimm-16} by considering allowed transitions $A$ which constrain the task space hypergraph $\mathcal{H}_{\mathcal{T}_\texttt{MRMP}}=(\tsv,\tse)$ such that a vertex $v_i\in\tsv$ iff $v_i.T_i=T_\texttt{COMPLETE}$.
This results in a $\mathcal{H}_{\mathcal{T}_\texttt{MRMP}}$ with a single vertex and no hyperarcs.
The motion hypergraph $\moh=(\mov,\moe)$ contains a pair of vertices, one for the start and one for the goal configuration in $\mathcal{W}_i$, along with the virtual source and target vertices.
A single move hyperarc $E^\texttt{move}_\mathcal{M}$ connects the start and goal configurations, and virtual hyperarcs connect these to the virtual source and target vertices.
As there are no meaningful transitions; the transition-extended hypergraph mirrors the motion hypergraph.
The computation of the path $E^\texttt{move}_\mathcal{M}.\pi$ provides a solution to the MRMP problem directly in the composite configuration space thus no conflict resolution stage is required.
}

\subsection{Decoupled Representation}
\begin{figure}[]
    \centering
    \includegraphics[width=.55\linewidth]{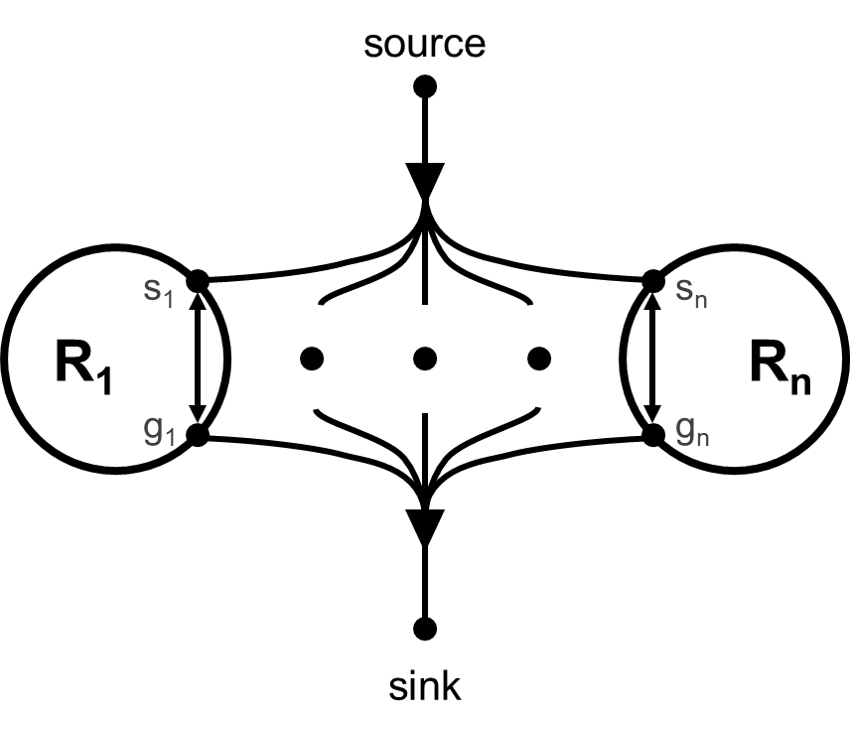}
    \caption{
    The motion hypergraph for a decoupled representation of the multi-robot motion planning problem contains single hyperarc within each individual robot subspace in addition to the source and sink hyperarcs. 
    The hyperarc within the task space element for each individual robot is an abstraction of the path across the corresponding roadmap within that element. 
    All robots must be included in the tail set of the sink hyperarc to capture the requirement that all robots reach their goal.}
    \label{fig:mrmp_grounded_hypergraph}
\end{figure}

\textcolor{black}{
MRMP-DaSH can be reduced to MRMP methods which utilize decoupled representations such as ~\cite{sl-uppccdpmrs-2002,smsa-rmmpucs-21} by considering allowed transitions $A$ which constrains $\mathcal{H}_{\mathcal{T}_\texttt{MRMP}}$ such that a vertex $v_i\in\tsv$ iff $|v_i.T_i.B_i| < |B|$ and $\tse=\emptyset$.
It is often further constrained such that each task space element represented in $\tsv$ contains only a single robot.
The corresponding motion hypergraph contains a start and goal vertex within each task space element represented by a vertex in $\tsv$ and a move hyperarc between them (Fig.~\ref{fig:mrmp_grounded_hypergraph}).
Additional hyperarcs connect the set of valid start and goal vertices to the virtual source and target vertices.
Fig.~\ref{fig:mrmp_grounded_hypergraph} depicts $\moh$ for a fully decoupled representation.
}

\textcolor{black}{
There are no meaningful transitions in $\moh$, so the transition-extended hypergraph $\teh$ resembles $\moh$.
The independent paths in the move hyperarcs are not guaranteed to be collision-free with each other, so the conflict resolution stage is required to produce a solution valid in $T_\texttt{COMPLETE}$.
The algorithm design choices for constructing the independent paths and resolving conflicts produce different MRMP algorithms with varying properties.
For example, using asymptotically optimal sampling-based approaches to iteratively construct and improve the individual paths can enable hybrid search methods such as CBS-MP~\cite{smsa-rmmpucs-21} to provide asymptotically optimal solutions.
Alternatively, priority-based methods such as Decoupled PRM~\cite{sl-uppccdpmrs-2002} or PBS~\cite{mhslk-swcpfmpf-19} applied to roadmaps will provide fast, suboptimal solutions.
The properties of grid world MAPF algorithms~\cite{ssfs-cbsomapf,mhslk-swcpfmpf-19} depend on the resolution of their grid representation.
}

\subsection{Hybrid Representation}

\textcolor{black}{
Hyrbid representations for MRMP-DaSH include composition hyperarcs in $\mathcal{H}_{\mathcal{T}_\texttt{MRMP}}$ allowing the subsequent representation layers to transition between $\cspace$ compositions.
This requires a decision to be made on when to change compositions.
Methods like M*~\cite{wc-sefmpp-15} change compositions as a means of resolving conflicts between decoupled paths.
This results in an iterative construction and search approach which lazily adds composition hyperarcs to the motion hypergraph $\moh$ around conflict locations.
The transition-extended hypergraph $\teh$ ensures that all solutions contain a single continuous path for each robot in the problem.
}

\section{Application to Multi-Manipulator Rearrangement}
\label{section:multi-manip}
\textcolor{black}{
We consider multi-manipulator rearrangement problems to illustrate our proposed approach and its benefits.
This application is referred to as MM-DaSH.
We first define the problem: Given a set of moveable bodies $\mathcal{B}=\mathcal{R}\cup\mathcal{O}$ consisting of robot manipulators $\mathcal{R}$ and manipulable objects $\mathcal{O}$, an action space expressed as allowed transitions $\mathcal{A}$, and an environment,
the planner must find a sequence of actions and motions for the robots to take such that the manipulable objects move from a start pose $p_\texttt{start}$ to a satisfactory goal pose $p_\texttt{goal}$.
}

\textcolor{black}{
The task space $\mathcal{T}_\texttt{manip}$ with elements $T_i=(B_i,\mathcal{W}_i,C_i)$ where $B_i$ contains some set of robots and objects, $\mathcal{W}_i$ denotes their $\cspace$, and $C_i$ may constrain an object $o_j\in B_i$ to be at rest on a stable surface or in a stable grasp of some robot $r_k\in B_i$ in addition to the standard collision avoidance constraints.
The set of admissible task space elements $\mathcal{T}^*$ consists exclusively of elements $T_i$ where $B_i=\mathcal{B}$ and $C_i$ must constrain every object to either a stable surface or grasp in additional to the full set of collision avoidance constraints.
The constraint set $C_\texttt{COMPLETE}$ in $T_\texttt{complete}$ contains all allowed grasps and stable surfaces for the objects and robots in $\mathcal{B}$.
}

\subsection{Composite Representation}
\textcolor{black}{
MM-DaSH can be configured to existing multi-robot rearrangement methods such as~\cite{sb-smrgbmgwcc-20} by considering a set of allowable transitions $A$ which constrains the task space hypergraph $\tsh=(\tsv,\tse)$ such that a vertex $v_i\in\tsv$ iff $v_i.T_i\in\mathcal{T}^*$.
As all task space elements represented as vertices in $\tsv$ contain all $r,o\in\mathcal{B}$, there are no composition hyperarcs in $\tsh$.
The action space available to the robots determines the transition hyperarcs $E_\mathcal{T}^\texttt{trans}$ included in $\tsh$.
A hyperarc $E_\mathcal{T}^\texttt{trans}\in\tse$ may include a set of simultaneous, independent actions transitioning between task space elements in $\mathcal{T}^*$, so long as each robot and object is only involved in single action
(e.g. Robot 1 picks object A while Robot 2 places object B).
As all these transition hyperarcs have a head and tail set of size one, $\tsh$ is a graph.
}

\textcolor{black}{
The motion hypergraph encodes paths for allowed transitions and motions within individual task space elements between transitions.
The transition-extended hypergraph encodes possible sequences of simultaneous actions and motions for the system to take.
As any path computed through a sequence of task space elements always considers all moveable bodies and collision constraints, no conflict resolution is needed.
}

\textcolor{black}{
Existing rearrangement methods like SMART~\cite{sb-smrgbmgwcc-20} can be modeled in this framework and often perform simultaneous construction and search.
They build an oracle for planning move hyperarcs $E_\mathcal{M}^\texttt{move}\in\moh.\moe$ (e.g. pre-compute PRM* for each robot) and then rely on heuristics to decide which hyperarc $E_\mathcal{T}^\texttt{trans}$ to attempt to sample and add to the motion hypergraph $\moh$.
They then consider taking the equivalent hyperarcs in the transition-extended hypergraph $\teh$ until $\teh$ contains a hyperpath corresponding to a sequence of moves and transitions which solve the task.
}

\textcolor{black}{
\subsection{Decoupled Representation}
MM-DaSH can  be configured to consider smaller task space elements with allowable transitions $A$ which constrain $\tsh=(\tsv,\tsh)$ such that $v_i\in\tsv$ iff the $B_i\in v_i.T_i$ cannot be partitioned without relaxing a constraint in $C_i$ other than inter-body collision avoidance.
For example, grasping constraints result in task space elements $T_i=(B_i,\mathcal{W}_i,C_i),B_i=\{o_j,r_k\}$ where robot $r_k$ is grasping object $o_j$ according to a constraint in $C_i$.
Task space elements with no grasping constraints contain either only a single object or robot.
}

\textcolor{black}{
A decoupled representation requires that no composition hyperarcs be included in $\tsh$.
Transition hyperarcs $E_\mathcal{T}^\texttt{trans}$ must be between sets of vertices in $\tsv$ and follow the rules described in Section~\ref{section:task_space_hypergraph}.
These correspond to independent actions included in the action space (e.g. pick, place, handoff) as depicted in Fig.~\ref{fig:task-space-hypergraph}.
}

\textcolor{black}{
The variations of representation construction and search presented in Section~\ref{section:method} produce decoupled or hybrid approaches with varying theoretical properties over this decoupled task space representation.
Hyperpaths in the transition-extended hypergraph $\teh$ correspond to unvalidated solutions in the set of admissible task space elements, so a conflict resolution stage is required.
Options for this stage are discussed in Section~\ref{section:conflict-resolution}.
We provide the decision choices for two decoupled representation options with hybrid search techniques in Section~\ref{section:experiments} which are included in our empirical evaluations.
}

\subsection{Hybrid Representation}
\textcolor{black}{
The hybrid representation for MM-DaSH applies the same modifications to the decoupled representation as the MRMP-DaSH hybrid representation.
Composition hyperarcs are allowed in $\tsh$ and the subsequent representation layers.
The decision remains on where to include composition changes.
Further investigation on this question is outside the scope of this work.
}

\section{Representation Analysis}
\label{section:representation_analysis}
\textcolor{black}{Having an explicit representation of the planning space enables incredibly informed generic heuristic based on the representation itself.
The composite graph-based representation provides more complete information than the decoupled hypergraph-based representation,
however,} in many applications, the graph representation of the task space derived from only considering the admissible task space elements is too large to ever use explicitly.
Existing methods which consider only the admissible composite elements instead use heuristics tailored to their problem to greedily search an implicit representation of the graph~\cite{sb-smrgbmgwcc-20}.
In contrast, if a hypergraph representation of decomposed task space elements is small enough to build,
it allows the use of more generic heuristics computed from the various hypergraph layers such as the cost-to-go heuristic discussed in Section~\ref{section:a-star}.

In this section, we compare representing the task space with the traditional composite graph representation to the proposed decoupled hypergraph representation for the multi-manipulator problems discussed in Section~\ref{section:multi-manip}.
We show that the size of the decoupled hypergraph representation scales well with the size of the problem and thus is able to be constructed and used in heuristics.

We consider the pick, place, handoff action space, and assume that all robots can reach each other to perform handoffs and all objects to perform pick/place actions.
This is an overestimate as it ignores reachability constraints
but serves as an informative upperbound on the space complexity as a planning method must reason over reachability before eliminating potential transitions.
Each robot can hold at most one object.
Each object can be held by at most one robot.
For simplicity, we ignore specific grasp constraints parameters or object poses and just consider allocation of objects to robots.

\subsection{Graph Representation}
We first consider a graph-based composite representation of the task space.
The moveable bodies of a task space element contain the full set of robots $n$ robots and $m$ objects.

\subsubsection{Total Number of Vertices}

Let $P_{\text{max}}=\texttt{min}(m,n)$ be the maximum number of objects that can be held in a scene with $n$ robots and $m$ objects.
Let $p\in\{0,...,P_{\text{max}}\}$ be the number of objects currently held in a given task space element's constraints $A^{n\text{x}m}$.
Then the number of unique sets of $p$ objects that are held in a given task space element is $m\choose p$.
The number of unique sets of $p$ robots holding those objects in a given task space element is $n\choose p$.
Considering all possible assignments of the held objects to the robots and the different values of $p$, the total number of elements $\mathcal{S}$ is given in (\ref{TotalStates}).

\begin{align}
\label{TotalStates}
	& \mathcal{S}(m,n) = \sum^{P_{\text{max}}}_{p=0}{m\choose p}{n\choose p} p!
\end{align}

\subsubsection{Total Number of Edges}
We first consider the set of available place actions.
Let $l\in\{0,..,p\}$ be the number of objects placed in a given transition.
Then the number of unique sets of objects placed in a given transition is $p\choose l$.
The total number of transitions just performing a place actions is given by (\ref{CountPlace}) where the 1 is subtracted for the case $l=0$ and no transition occurs.

\begin{align}
\label{CountPlace}
	& -1 + \sum^{p}_{l=0}{p\choose l}
\end{align}

$n-p$ is the number of free robots in a given task space element.
$H_{\text{max}}=\texttt{min}(p-l,n-p)\}$ denotes the maximum number of objects handed off in a given transition.
Let $h\in\{0,...,H_{\text{max}}\}$ be the number of objects handed off in a given transition.
Then the number of unique sets of objects handed off in a given transition is ${p-l}\choose h$,
and the number of unique robots receiving the handoffs is ${n-p}\choose h$.
Considering all possible assignments, the total number of place and handoff transitions is given by (\ref{CountPlaceHandoff}) where the 1 is subtracted for the case $l=0,h=0$ and no transition occurs.

\begin{align}
\label{CountPlaceHandoff}
	& -1 + \sum^{p}_{l=0}{p\choose l}\sum^{H_\text{max}}_{h=0}{{p-l}\choose h}{{n-p}\choose h}h!
\end{align}

$m-p$ is the number of free objects in a given task space element.
$n-p-h$ denotes the number of robots not holding an object in an element and not receiving an object via a handoff in a given transition.
$G_\text{max}=\texttt{min}(m-p,n-p-h)$ is the maximum number of objects picked in a given transition.
Let $g\in\{0,G_\text{max}\}$ be the number of objects picked in a given transition.
Then the number of unique sets of objects being picked is ${m-p}\choose g$, 
and the the number of unique sets of robots picking is ${n-p-h}\choose g$.
Considering all possible assignments, the total number of transitions from a given task space element performing a set of handoff actions is given by (\ref{CountPickPlaceHandoff}) where the 1 is subtracted for the case $l=0,h=0,g=0$ and no transition occurs.

\begin{align}
\label{CountPickPlaceHandoff}
	& \mathcal{G}(m,n,p,h) = -1 + \sum^{G_\text{max}}_{g=0}{{m-p}\choose g}{{n-p-h}\choose g}g!
\end{align}

Let $\mathcal{H}$ be the total number of handoff and grasp action combinations possible given $m,n,p,$ and $l$,
and let $\mathcal{L}$ be the total number of transitions possible consisting of a combination of place, handoff, and pick actions given $m,n,$ and $p$.

\begin{align}
\label{CountAllHandoff}
	& \mathcal{H}(m,n,p,l) = \sum^{H_\text{max}}_{h=0}{{p-l}\choose h}{{n-p}\choose h}h!\mathcal{G}(m,n,p,h)\\
	& \mathcal{L}(m,n,p) =  \sum^{p}_{l=0}{p\choose l}\mathcal{H}(m,n,p,l)
\end{align}

The total number of edges $\mathcal{T}$ in the graph is given by the total number of edges from each of the vertices (\ref{TotalEdges}).
\begin{align}
\label{TotalEdges}
	& \mathcal{T}(m,n) = \sum^{P_{\text{max}}}_{p=0}{m\choose p}{n\choose p} p!\mathcal{L}(m,n,p)
\end{align}

\subsection{Hypergraph Representation}
\label{section:hypergraph_analysis}
The hypergraph-based representation of the decoupled task space considers decomposed task space elements containing every pairing of a robot grasping an object, each robot not holding an object, and each object not being grasped for a total of $mn + m + n$ vertices.

Each pick action is a transition from a robot only vertex and an object only vertex to the vertex for the task space element where the robot grasps that object.
A place action is this transition in reverse.
This results in $2mn$ transition hyperarcs for all pick/place actions.

Each handoff action is a transition from a pair of vertices where robot 1 is grasping the object and robot 2 is free to a pair of vertices where robot 1 is free and robot 2 is grasping the object.
Each of the $mn$ grasping vertices can be paired with any of the $n-1$ free robot vertices for a total of $mn^2 - mn$ handoff hyperarcs.
This results in a total of $mn^2 + mn$ hyperarcs to capture all possible transitions. 

\subsection{Comparison}

From (~\ref{TotalStates}), the graph-based representation maintains linear growth in the number of vertices when considering either a single robot or a single object, but exhibits exponential growth as soon there are both multiple robots and multiple objects.

In contrast, in the hypergraph representation, the number of vertices scales linearly with the number of robots while holding the number of objects constant, regardless of the number of objects. 
The same is true if the number of objects changes and the number of robots is held constant.

The number of transitions follows a similar trend.
If either there is either a single object or a single robot then the number of transitions is equal for graph- and hypergraph-based representations when the other quantity (robots or objects) increases.
As soon as a second object or a second robot is introduced, the growth of the number of task space transitions (edges) in the graph-based representation becomes exponential.
In the hypergraph-based representation, the number of hyperarcs scales quadratically with the number of robots and linearly with the number of objects.
This follows directly from the total number of hyperarcs discussed in Section~\ref{section:hypergraph_analysis}.

\textcolor{black}{
Despite containing more information, the exponential growth in the size of the composite graph-based representation prevents it from being utilized in planning.
The more concise hypergraph-based representation can be both constructed and leveraged to inform heuristics which are not problem specific.
We present one instance of this in Section~\ref{section:a-star} where heuristic values come from simple paths in the motion hypergraph.
}

\textcolor{black}{
To provide an intuitive visualization of the difference in representation size, we depict the graph and hypergraph representation for 2 robots and 4 objects along with a table showing the immediate size difference in smaller problems (Fig.~\ref{fig:intuitive_graphs}).
}

\subsection{Alternative Problem Spaces}
\textcolor{black}{
To examine the generality of the DaSH framework, we consider the size of the hypergraph-based representation for alternative problem spaces.
}

\subsubsection{Two Robot Grasping}
\label{section:two-robot-grasping}
\textcolor{black}{
We first consider a problem space where each object requires two robots to be simultaneous grasping an object to move it.
We assume that each robot involved in a grasp has a specific role (i.e. robot 1 and robot 2).
We maintain the $m+n$ vertices for each object alone and each manipulator alone task space element.
Instead of the $mn$ vertices corresponding to task space elements for object-robot pairs, we have $mn^2-mn$ elements for each permutation of two robots and one object for a total of $mn^2 + m + n$ vertices in the task space hypergraph.
}

\textcolor{black}{
In this problem, there are $2mn^2-2mn$ pick/place transitions for every combination of robot pairs grasping an object.
Additionally, there are $mn(n-1)(n-2)(n-3)$ handoff transitions for as each of the $m$ objects can be handed off between every ordered pair of robots.
Thus the number of transitions in the hypergraph-based representation still scales linearly with the number of objects and quartically with the number of robots.
}

\textcolor{black}{
The number of vertices in the task space hypergraph can be modeled for larger sets of $k$ robots required for grasping an object by considering $m\frac{n!}{(n-k)!}$ task space elements where $k$ robots grasp each object with specific roles in the grasp in addition to the isolated robot and object element vertices.
Pick/place transitions is always twice this number for moving in either direction.
Meanwhile, there are $m\frac{n!}{(n-k)!}\frac{(n-k)!}{(n-2k)!}$ handoff transitions.
}

\subsubsection{Stacking}
\textcolor{black}{
We consider a problem space where a set of $m$ objects must be stacked in a particular order and can only be stacked in that order.
Each object can be grasped by at most one robot at a time.
From our original pick, place, handoff action space,
we add an additional $m$ task space elements corresponding to every partial stack of $i\in[1,m]$ objects each of which corresponds to an additional vertex in the task space hypergraph.
The addition of each of the $m$ objects to the stack may be done by any of the $n$ robot for an additional $mn$ transitions.
Thus the growth of the representation size is maintains the same asymptotic behavior.
}

\textcolor{black}{
Now consider instead the stacking problem, but where each object requires two robots to grasp it as described in Section~\ref{section:two-robot-grasping}.
We still have the $m$ additional vertices for each partial stack.
Instead of the $mn$ stacking transitions in the single robot grasp version, we now have $mn(n-1)$ stacking transitions.
The asymptotic growth of the representation size again remains unchanged from the two robot pick, place, handoff rearrangement problem.
}

\subsubsection{Supported Stacking}
\textcolor{black}{
We add an additional constraint to the stacking problem, where one robot must be \textit{supporting} the stack before an object can be added.
This adds an additional $mn$ vertices corresponding to task space elements where each of the $n$ robots is supporting one of the $m$ partial stacks.
Additionally, instead of $mn$ stacking transitions for the single robot grasp problem or $mn(n-1)$ stacking transitions for the two robot grasp problem,
there are now $mn(n-1)$ and $mn(n-1)(n-2)$ supported stacking transitions respectively, where each stacking action now involves an additional robot supporting the stack.
In either case, this is less than or equal to the number of handoff transitions in either problem's task space hypergraph, leaving the asymptotic growth unchanged.
}

\section{Experimental Evaluation}
\label{section:experiments}

\textcolor{black}{
In our experiments, we demonstrate the improvement in planning times from using the hypergraph-based representation for both problems with large numbers of objects and geometrically constrained manipulation tasks.
To do this, we evaluate two decoupled representation configurations of the approach for the multi-manipulator rearrangement problem defined in Section~\ref{section:multi-manip}, MM-DaSH and Greedy-MM-DaSH, on a set of physical and simulated multi-robot manipulation problems depicted in Figures~\ref{fig:physical_image},~\ref{fig:shelves_image}.
We compare against SMART~\cite{sb-smrgbmgwcc-20} as a multi-manipulator method considering a composite representation for the rearrangement tasks.
In all scenarios, the objective is to move the given objects from their respective start locations to specified goal locations.
Robots can perform pick, place, and handoff actions.
We report the time to return the first solution and the cost of that solution for each method.
Any planning attempt longer than $10^4$ seconds is considered to have failed.
}

\subsection{Methods}

\textcolor{black}{
We describe the two DaSH configurations and the implementation of SMART~\cite{sb-smrgbmgwcc-20}.
}

\subsubsection{MM-DaSH}

\textcolor{black}{
We describe the \textit{construction, task planning}, and \textit{conflict resolution} used in both MM-DaSH and Greedy-MM-DaSH.
}

\textcolor{black}{
\textbf{Construction-}
Both variants construct the entire task space hypergraph $\tsh$ in the initial representation construction.
The size relative the number of robots and objects is given in Section~\ref{section:representation_analysis}.
}

\textcolor{black}{
Each round of representation expansion samples one additional motion transition for each task space transition.
We use an IK-based sampling method for grasps and handoffs.
A small roadmap is initialized with 10 vertices for each task space element using PRM*.
The move planner uses these roadmaps to compute the paths included in the move hyperarcs by connecting the transition path start/end points and querying paths over the roadmaps.
Each round of representation expansion adds 10 additional samples to each roadmap.
}

\textcolor{black}{
\textbf{Task Planning-}
MM-DaSH uses NBS and the A*-like hyperpath query described in Section~\ref{section:a-star} to find optimal solutions for the current representation.
This can require many calls to both the task planning and conflict resolution layers.
}

\textcolor{black}{
Greedy-MM-DaSH uses the Greedy hyperpath query described in Section~\ref{section:greedy-search} and expands the representation after a single attempt at task planning and conflict resolution.
}

\textcolor{black}{
\textbf{Conflict Resolution-}
Both methods use the scheduled variant of CBS-MP~\cite{smsa-rmmpucs-21} to resolve conflicts in the optimistic schedule produced by the task planning layer.
}

\textcolor{black}{
\textbf{Solution-}
Both methods return a solution once the representation contains one.
MM-DaSH will find the optimal solution for that representation while Greedy-MM-DaSH will return the first solution it finds.
}

\subsubsection{SMART}

\textcolor{black}{
We describe the details of our implementation of SMART~\cite{sb-smrgbmgwcc-20}.
}

\textcolor{black}{
\textbf{Construction-}
We first create the object centric mode graph used to guide planning in SMART.
We then run PRM* to create an initial roadmap with 100 vertices for each robot.
We found that initializing each roadmap with 100 vertices led to the highest success rate when using the dRRT*~\cite{ssdhb-dsaiammp-20} style multi-robot motion planning in later stages of the algorithm as detailed in~\cite{sb-smrgbmgwcc-20}.
The need for denser roadmaps than the CBS-MP style motion planning is an interesting question, but is outside the scope of this work.
}

\textcolor{black}{
We modified the implementation of SMART to sample transition (picks, places, handoffs) during the construction to create a fair comparison to the DaSH variants which also sample transition in the construction phase.
These are connected to the PRM* generated roadmaps.
}

\textcolor{black}{
\textbf{Search-}
The implementation of the search stage follows the description in the original manuscript~\cite{sb-smrgbmgwcc-20}.
We use the same MAPF formulation over the object centric mode graph as the guiding heuristic.
The goal bias was 0.9.
This was experimentally found to produce the best results.
The same dRRT* style motion planning was used within each mode.
}

\textcolor{black}{
\textbf{Solution-}
We return the first valid solution SMART finds.
}

\textcolor{black}{
Both MM-DaSH variants and SMART~\cite{sb-smrgbmgwcc-20} were implemented in C++ in the Parasol Planning Library.
We report the total search time in Fig.~\ref{fig:sorting_results},~\ref{fig:shelves}
and construction time, success rate, and solution quality in Tables~\ref{tab:sorting_results}.
}

\subsection{Experiment Configuration}
We evaluate the methods on a set of rearrangement tasks.
A \textit{sorting} scenario highlights the ability of the methods to plan for increasing numbers of robots and objects.
\textcolor{black}{
A \textit{shelving} scenario demonstrates the ability of the MM-DaSH variants to account for geometric constraints arising from object placement.
}
Each method was run with 10 random seeds in each scenario.
Experiments were run using a desktop computer with an Intel Core i9-10900KF CPU at 3.7 GHz, 128 GB of RAM.

\textcolor{black}{
We compute plans for virtual robots in each scenario.}
All robots are UR5e manipulators with hand-e grippers.
A physical demonstration of the two robot sorting scenario can be seen in the video linked in Fig.~\ref{fig:physical_image}.

SMART was adapted to account for transition paths to perform task space element switches instead of single configuration switches to address the added complexity of using a two finger gripper instead of the vacuum style gripper used in the original SMART experiments.

\subsection{Sorting}
\textcolor{black}{
To evaluate the ability of the method to efficiently find plans for large numbers of objects,}
we consider a scenario which imitates sorting tasks that may be encountered in warehouses or factories.
A set of objects lies in a common workspace for a set of robots.
Each object belongs to a particular class.
The team of robots must sort the objects from the mixed group in the common workspace into bins of the corresponding class.
In a full industry application, a computer vision component would be utilized to identify object classes and poses, but we focus only on the planning portion here, and assume knowledge of this information.
Our objects are simple cubes, and the class is denoted by the color in Figure~\ref{fig:physical_image}.

\begin{figure*}
    \centering
    \centering
     \subfloat[Sorting - 2 Robots - Cross]{
    	\includegraphics[height=.3\linewidth]{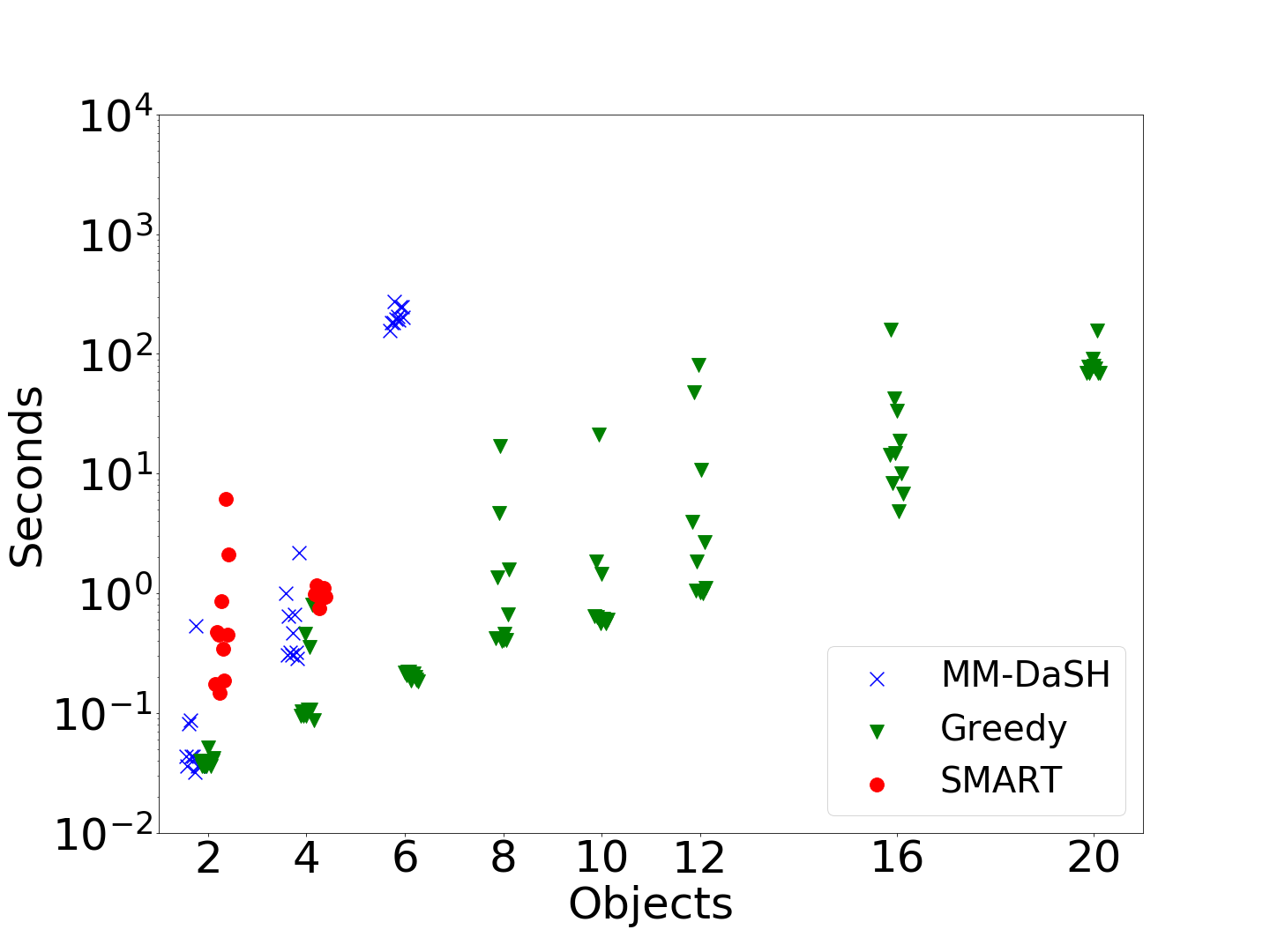}
    	\label{fig:small_sorting_cross}
    }
     \subfloat[Sorting - 2 Robots - Random]{
    	\includegraphics[height=.3\linewidth]{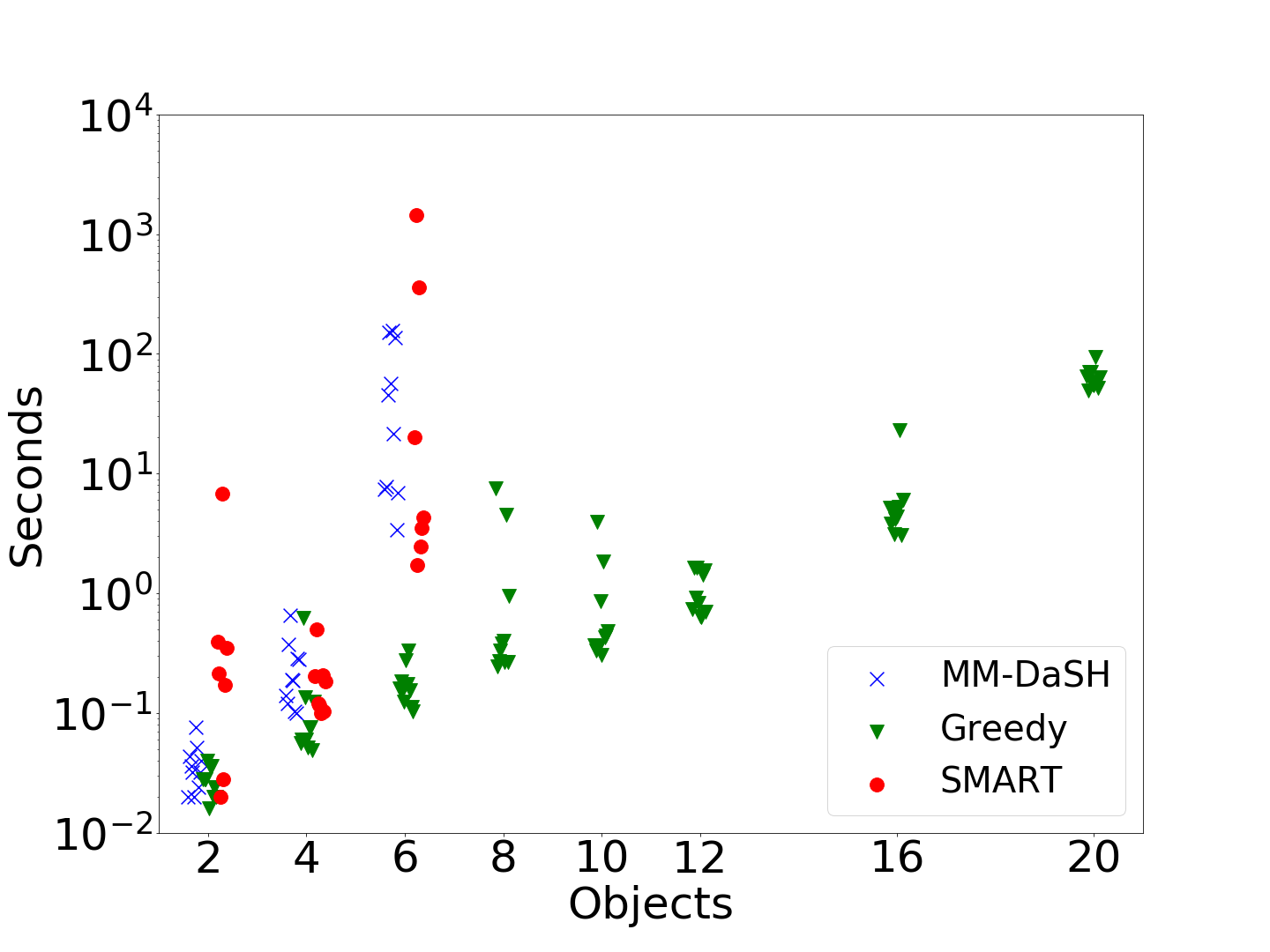}
    	\label{fig:small_sorting_rand}
    }\\
     \subfloat[Sorting - 4 Robots - Cross]{
    	\includegraphics[height=.3\linewidth]{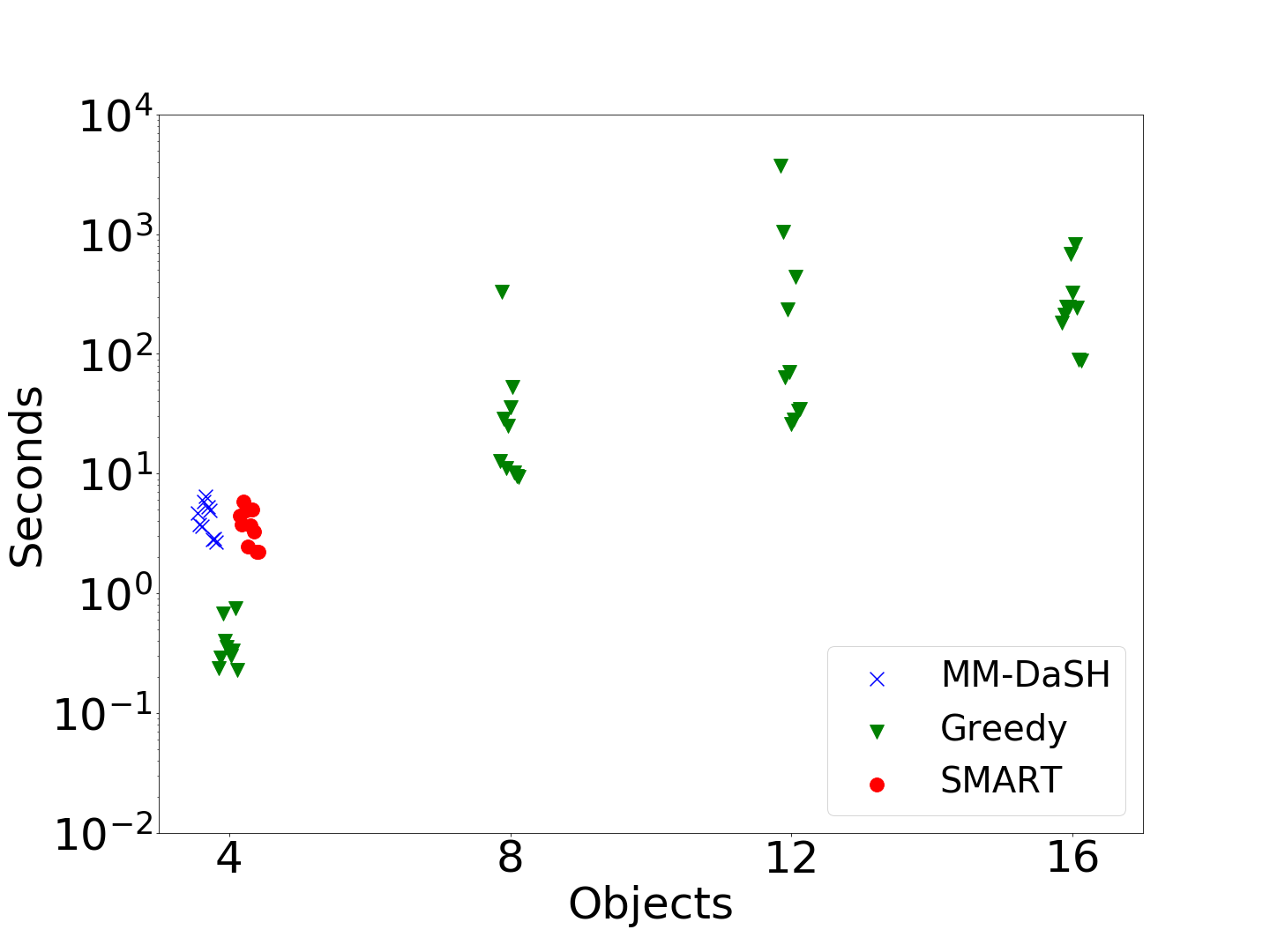}
    	\label{fig:big_sorting_cross}
    }
     \subfloat[Sorting - 4 Robots - Random]{
    	\includegraphics[height=.3\linewidth]{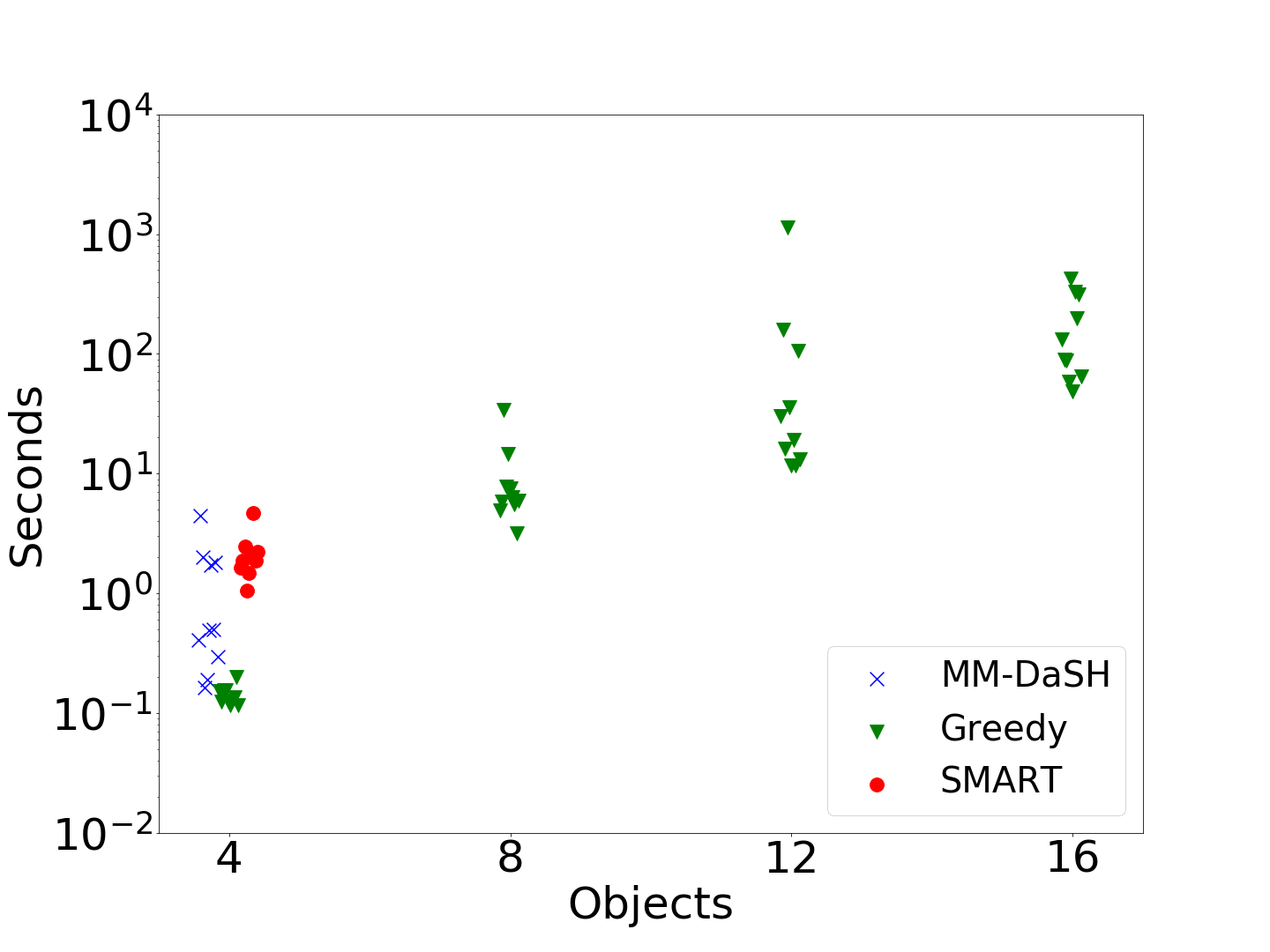}
    	\label{fig:big_sorting_rand}
    }
    \caption{
    Each figure shows the search time for every seed.
    The Y-axis denotes the seconds taken to compute a plan.
    The X-axis denotes the number of objects in the problem.
    The asymptotically optimal version, MM-DaSH, is marked with blue X's.
    The greedy variant, Greedy-MM-DaSH, is marked with green triangles.
    The composite version, SMART~\cite{sb-smrgbmgwcc-20}, is marked with red circles.
    Figures (a,b) show the results for 2 robot sorting scenarios.
    Figures (c,d) show the results for 4 robot sorting scenarios.
    Figures (a,c) have start locations for each object such that it must be handed off between robots to reach the goal location.
    Figures (b,d) have random start locations for each object.
    }
    \label{fig:sorting_results}
\end{figure*}

\begin{table*}[]
\centering
\caption{Detailed results for both two and four robot sorting scenarios.}
\begin{tabular}{|c|c|c|c|rr|rr|rr|r|}
\hline
\multirow{2}{*}{Robots} & \multirow{2}{*}{Objects} & \multirow{2}{*}{Scenario} & \multirow{2}{*}{Method}    & \multicolumn{2}{c|}{Construction (s)}                                & \multicolumn{2}{c|}{Search (s)}                                      & \multicolumn{2}{c|}{\begin{tabular}[c]{@{}c@{}}Cost \\ (Timesteps)\end{tabular}} & \multicolumn{1}{c|}{\multirow{2}{*}{Success}} \\ \cline{5-10}
                        &                          &                           &                            & \multicolumn{1}{c|}{Avg}              & \multicolumn{1}{c|}{Std Dev} & \multicolumn{1}{c|}{Avg}              & \multicolumn{1}{c|}{Std Dev} & \multicolumn{1}{c|}{Avg}                     & \multicolumn{1}{l|}{Std Dev}      & \multicolumn{1}{c|}{}                         \\ \hline
\multirow{27}{*}{2}     & \multirow{6}{*}{2}       & \multirow{3}{*}{Cross}    & \cellcolor[HTML]{EFEFEF}MM-DaSH                    & \multicolumn{1}{r|} {\textbf{\cellcolor[HTML]{EFEFEF}11.39}}   &\cellcolor[HTML]{EFEFEF} 2.41                         & \multicolumn{1}{r|}{\cellcolor[HTML]{EFEFEF}0.10}             & \cellcolor[HTML]{EFEFEF}0.15                         & \multicolumn{1}{r|}{\textbf{\cellcolor[HTML]{EFEFEF}993.40}}         & \cellcolor[HTML]{EFEFEF}171.96                            & \textbf{\cellcolor[HTML]{EFEFEF}1.00}                                 \\ \cline{4-11} 
                        &                          &                           & Greedy                     & \multicolumn{1}{r|}{\textbf{11.39}}   & 2.41                         & \multicolumn{1}{r|}{\textbf{0.04}}    & 0.01                         & \multicolumn{1}{r|}{1000.80}                 & 172.15                            & \textbf{1.00}                                 \\ \cline{4-11} 
                        &                          &                           & \cellcolor[HTML]{EFEFEF}SMART                      & \multicolumn{1}{r|}{\cellcolor[HTML]{EFEFEF}156.96}           & \cellcolor[HTML]{EFEFEF}5.75                         & \multicolumn{1}{r|}{\cellcolor[HTML]{EFEFEF}1.13}             & \cellcolor[HTML]{EFEFEF}1.85                         & \multicolumn{1}{r|}{\cellcolor[HTML]{EFEFEF}14139.00}                & \cellcolor[HTML]{EFEFEF}23648.27                          & \textbf{\cellcolor[HTML]{EFEFEF}1.00}                                 \\ \cline{3-11} 
                        &                          & \multirow{3}{*}{Random}   & MM-DaSH                    & \multicolumn{1}{r|}{\textbf{12.01}}   & 2.81                         & \multicolumn{1}{r|}{0.04}             & 0.02                         & \multicolumn{1}{r|}{\textbf{641.20}}         & 350.86                            & \textbf{1.00}                                 \\ \cline{4-11} 
                        &                          &                           & \cellcolor[HTML]{EFEFEF}Greedy                     & \multicolumn{1}{r|}{\textbf{\cellcolor[HTML]{EFEFEF}12.01}}   & \cellcolor[HTML]{EFEFEF}2.81                         & \multicolumn{1}{r|}{\textbf{\cellcolor[HTML]{EFEFEF}0.03}}    & \cellcolor[HTML]{EFEFEF}0.01                         & \multicolumn{1}{r|}{\cellcolor[HTML]{EFEFEF}664.40}                  & \cellcolor[HTML]{EFEFEF}352.07                            & \textbf{\cellcolor[HTML]{EFEFEF}1.00}                                 \\ \cline{4-11} 
                        &                          &                           & SMART                      & \multicolumn{1}{r|}{159.71}           & 8.78                         & \multicolumn{1}{r|}{1.14}             & 2.50                         & \multicolumn{1}{r|}{3875.71}                 & 4257.81                           & 0.70                                          \\ \cline{2-11} 
                        & \multirow{6}{*}{4}       & \multirow{3}{*}{Cross}    & \cellcolor[HTML]{EFEFEF}MM-DaSH                    & \multicolumn{1}{r|}{\textbf{\cellcolor[HTML]{EFEFEF}23.32}}   & \cellcolor[HTML]{EFEFEF}7.18                         & \multicolumn{1}{r|}{\cellcolor[HTML]{EFEFEF}0.65}             & \cellcolor[HTML]{EFEFEF}0.59                         & \multicolumn{1}{r|}{\textbf{\cellcolor[HTML]{EFEFEF}1845.60}}        & \cellcolor[HTML]{EFEFEF}191.95                            & \textbf{\cellcolor[HTML]{EFEFEF}1.00}                                 \\ \cline{4-11} 
                        &                          &                           & Greedy                     & \multicolumn{1}{r|}{\textbf{23.32}}   & 7.18                         & \multicolumn{1}{r|}{\textbf{0.23}}    & 0.24                         & \multicolumn{1}{r|}{2035.10}                 & 290.85                            & \textbf{1.00}                                 \\ \cline{4-11} 
                        &                          &                           & \cellcolor[HTML]{EFEFEF}SMART                      & \multicolumn{1}{r|}{\cellcolor[HTML]{EFEFEF}167.48}           & \cellcolor[HTML]{EFEFEF}17.52                        & \multicolumn{1}{r|}{\cellcolor[HTML]{EFEFEF}0.97}             & \cellcolor[HTML]{EFEFEF}0.13                         & \multicolumn{1}{r|}{\cellcolor[HTML]{EFEFEF}4387.25}                 & \cellcolor[HTML]{EFEFEF}1312.74                           & \cellcolor[HTML]{EFEFEF}0.80                                          \\ \cline{3-11} 
                        &                          & \multirow{3}{*}{Random}   & MM-DaSH                    & \multicolumn{1}{r|}{\textbf{24.82}}   & 8.42                         & \multicolumn{1}{r|}{0.24}             & 0.17                         & \multicolumn{1}{r|}{\textbf{1059.40}}        & 380.99                            & \textbf{1.00}                                 \\ \cline{4-11} 
                        &                          &                           & \cellcolor[HTML]{EFEFEF}Greedy                     & \multicolumn{1}{r|}{\textbf{\cellcolor[HTML]{EFEFEF}24.82}}   & \cellcolor[HTML]{EFEFEF}8.42                         & \multicolumn{1}{r|}{\textbf{\cellcolor[HTML]{EFEFEF}0.13}}    & \cellcolor[HTML]{EFEFEF}0.18                         & \multicolumn{1}{r|}{\cellcolor[HTML]{EFEFEF}1199.80}                 & \cellcolor[HTML]{EFEFEF}395.14                            & \textbf{\cellcolor[HTML]{EFEFEF}1.00}                                 \\ \cline{4-11} 
                        &                          &                           & SMART                      & \multicolumn{1}{r|}{157.36}           & 7.04                         & \multicolumn{1}{r|}{0.19}             & 0.13                         & \multicolumn{1}{r|}{1920.50}                 & 1361.05                           & 0.80                                          \\ \cline{2-11} 
                        & \multirow{5}{*}{6}       & \multirow{2}{*}{Cross}    & \cellcolor[HTML]{EFEFEF}MM-DaSH                    & \multicolumn{1}{r|}{\textbf{\cellcolor[HTML]{EFEFEF}40.74}}   & \cellcolor[HTML]{EFEFEF}11.81                        & \multicolumn{1}{r|}{\cellcolor[HTML]{EFEFEF}207.75}           & \cellcolor[HTML]{EFEFEF}36.21                        & \multicolumn{1}{r|}{\textbf{\cellcolor[HTML]{EFEFEF}2399.20}}        & \cellcolor[HTML]{EFEFEF}234.32                            & \textbf{\cellcolor[HTML]{EFEFEF}1.00}                                 \\ \cline{4-11} 
                        &                          &                           & Greedy                     & \multicolumn{1}{r|}{\textbf{40.74}}   & 11.81                        & \multicolumn{1}{r|}{\textbf{0.21}}    & 0.02                         & \multicolumn{1}{r|}{2752.80}                 & 320.78                            & \textbf{1.00}                                 \\ \cline{3-11} 
                        &                          & \multirow{3}{*}{Random}   & \cellcolor[HTML]{EFEFEF}MM-DaSH                    & \multicolumn{1}{r|}{\textbf{\cellcolor[HTML]{EFEFEF}44.90}}   & \cellcolor[HTML]{EFEFEF}21.09                        & \multicolumn{1}{r|}{\cellcolor[HTML]{EFEFEF}59.26}            & \cellcolor[HTML]{EFEFEF}63.77                        & \multicolumn{1}{r|}{\textbf{\cellcolor[HTML]{EFEFEF}1607.30}}        & \cellcolor[HTML]{EFEFEF}546.54                            & \textbf{\cellcolor[HTML]{EFEFEF}1.00}                                 \\ \cline{4-11} 
                        &                          &                           & Greedy                     & \multicolumn{1}{r|}{\textbf{44.90}}   & 21.09                        & \multicolumn{1}{r|}{\textbf{0.18}}    & 0.07                         & \multicolumn{1}{r|}{1969.70}                 & 640.94                            & \textbf{1.00}                                 \\ \cline{4-11} 
                        &                          &                           & \multicolumn{1}{l|}{\cellcolor[HTML]{EFEFEF}SMART} & \multicolumn{1}{r|}{\cellcolor[HTML]{EFEFEF}289.99}           & \cellcolor[HTML]{EFEFEF}86.01                        & \multicolumn{1}{r|}{\cellcolor[HTML]{EFEFEF}260.48}           & \cellcolor[HTML]{EFEFEF}532.82                       & \multicolumn{1}{r|}{\cellcolor[HTML]{EFEFEF}7911.29}                 & \cellcolor[HTML]{EFEFEF}4389.18                           & \cellcolor[HTML]{EFEFEF}0.70                                          \\ \cline{2-11} 
                        & \multirow{2}{*}{8}       & Cross                     & Greedy                     & \multicolumn{1}{r|}{\textbf{122.55}}  & 172.79                       & \multicolumn{1}{r|}{\textbf{2.73}}    & 5.17                         & \multicolumn{1}{r|}{\textbf{3758.30}}        & 254.71                            & \textbf{1.00}                                 \\ \cline{3-11} 
                        &                          & Random                    & \cellcolor[HTML]{EFEFEF}Greedy                     & \multicolumn{1}{r|}{\textbf{\cellcolor[HTML]{EFEFEF}74.24}}   & \cellcolor[HTML]{EFEFEF}23.53                        & \multicolumn{1}{r|}{\textbf{\cellcolor[HTML]{EFEFEF}1.53}}    & \cellcolor[HTML]{EFEFEF}2.51                         & \multicolumn{1}{r|}{\textbf{\cellcolor[HTML]{EFEFEF}2584.60}}        & \cellcolor[HTML]{EFEFEF}492.95                            & \textbf{\cellcolor[HTML]{EFEFEF}1.00}                                 \\ \cline{2-11} 
                        & \multirow{2}{*}{10}      & Cross                     & Greedy                     & \multicolumn{1}{r|}{\textbf{89.92}}   & 36.93                        & \multicolumn{1}{r|}{\textbf{2.87}}    & 6.46                         & \multicolumn{1}{r|}{\textbf{4612.30}}        & 266.45                            & \textbf{1.00}                                 \\ \cline{3-11} 
                        &                          & Random                    & \cellcolor[HTML]{EFEFEF}Greedy                     & \multicolumn{1}{r|}{\textbf{\cellcolor[HTML]{EFEFEF}105.59}}  & \cellcolor[HTML]{EFEFEF}63.35                        & \multicolumn{1}{r|}{\textbf{\cellcolor[HTML]{EFEFEF}0.94}}    & \cellcolor[HTML]{EFEFEF}1.15                         & \multicolumn{1}{r|}{\textbf{\cellcolor[HTML]{EFEFEF}3092.10}}        & \cellcolor[HTML]{EFEFEF}553.30                            & \textbf{\cellcolor[HTML]{EFEFEF}1.00}                                 \\ \cline{2-11} 
                        & \multirow{2}{*}{12}      & Cross                     & Greedy                     & \multicolumn{1}{r|}{\textbf{163.50}}  & 100.39                       & \multicolumn{1}{r|}{\textbf{15.16}}   & 27.13                        & \multicolumn{1}{r|}{\textbf{5802.80}}        & 450.26                            & \textbf{1.00}                                 \\ \cline{3-11} 
                        &                          & Random                    & \cellcolor[HTML]{EFEFEF}Greedy                     & \multicolumn{1}{r|}{\textbf{\cellcolor[HTML]{EFEFEF}170.32}}  & \cellcolor[HTML]{EFEFEF}111.31                       & \multicolumn{1}{r|}{\textbf{\cellcolor[HTML]{EFEFEF}1.08}}    & \cellcolor[HTML]{EFEFEF}0.44                         & \multicolumn{1}{r|}{\textbf{\cellcolor[HTML]{EFEFEF}3880.90}}        & \cellcolor[HTML]{EFEFEF}420.81                            & \textbf{\cellcolor[HTML]{EFEFEF}1.00}                                 \\ \cline{2-11} 
                        & \multirow{2}{*}{16}      & Cross                     & Greedy                     & \multicolumn{1}{r|}{\textbf{450.36}}  & 470.13                       & \multicolumn{1}{r|}{\textbf{31.27}}   & 46.36                        & \multicolumn{1}{r|}{\textbf{7579.80}}        & 486.61                            & \textbf{1.00}                                 \\ \cline{3-11} 
                        &                          & Random                    & \cellcolor[HTML]{EFEFEF}Greedy                     & \multicolumn{1}{r|}{\textbf{\cellcolor[HTML]{EFEFEF}263.56}}  & \cellcolor[HTML]{EFEFEF}103.63                       & \multicolumn{1}{r|}{\textbf{\cellcolor[HTML]{EFEFEF}6.31}}    & \cellcolor[HTML]{EFEFEF}5.97                         & \multicolumn{1}{r|}{\textbf{\cellcolor[HTML]{EFEFEF}4884.40}}        & \cellcolor[HTML]{EFEFEF}866.13                            & \textbf{\cellcolor[HTML]{EFEFEF}1.00}                                 \\ \cline{2-11} 
                        & \multirow{2}{*}{20}      & Cross                     & Greedy                     & \multicolumn{1}{r|}{\textbf{484.06}}  & 134.07                       & \multicolumn{1}{r|}{\textbf{1767.64}} & 5323.53                      & \multicolumn{1}{r|}{\textbf{9357.20}}        & 463.72                            & \textbf{1.00}                                 \\ \cline{3-11} 
                        &                          & Random                    & \cellcolor[HTML]{EFEFEF}Greedy                     & \multicolumn{1}{r|}{\textbf{\cellcolor[HTML]{EFEFEF}651.83}}  & \cellcolor[HTML]{EFEFEF}248.32                       & \multicolumn{1}{r|}{\textbf{\cellcolor[HTML]{EFEFEF}63.50}}   & \cellcolor[HTML]{EFEFEF}13.08                        & \multicolumn{1}{r|}{\textbf{\cellcolor[HTML]{EFEFEF}6467.50}}        & \cellcolor[HTML]{EFEFEF}882.46                            & \textbf{\cellcolor[HTML]{EFEFEF}1.00}                                 \\ \hline
\multirow{12}{*}{4}     & \multirow{6}{*}{4}       & \multirow{3}{*}{Cross}    & MM-DaSH                    & \multicolumn{1}{r|}{\textbf{43.45}}   & 9.67                         & \multicolumn{1}{r|}{4.29}             & 1.35                         & \multicolumn{1}{r|}{\textbf{1946.50}}        & 175.20                            & \textbf{1.00}                                 \\ \cline{4-11} 
                        &                          &                           & \cellcolor[HTML]{EFEFEF}Greedy                     & \multicolumn{1}{r|}{\textbf{\cellcolor[HTML]{EFEFEF}43.45}}   & \cellcolor[HTML]{EFEFEF}9.67                         & \multicolumn{1}{r|}{\textbf{\cellcolor[HTML]{EFEFEF}0.39}}    & \cellcolor[HTML]{EFEFEF}0.18                         & \multicolumn{1}{r|}{\cellcolor[HTML]{EFEFEF}3277.20}                 & \cellcolor[HTML]{EFEFEF}344.71                            & \textbf{\cellcolor[HTML]{EFEFEF}1.00}                                 \\ \cline{4-11} 
                        &                          &                           & SMART                      & \multicolumn{1}{r|}{341.01}           & 167.41                       & \multicolumn{1}{r|}{3.79}             & 1.27                         & \multicolumn{1}{r|}{5486.60}                 & 1484.64                           & \textbf{1.00}                                 \\ \cline{3-11} 
                        &                          & \multirow{3}{*}{Random}   & \cellcolor[HTML]{EFEFEF}MM-DaSH                    & \multicolumn{1}{r|}{\textbf{\cellcolor[HTML]{EFEFEF}36.15}}   & \cellcolor[HTML]{EFEFEF}8.88                         & \multicolumn{1}{r|}{\cellcolor[HTML]{EFEFEF}1.21}             & \cellcolor[HTML]{EFEFEF}1.35                         & \multicolumn{1}{r|}{\textbf{\cellcolor[HTML]{EFEFEF}1269.80}}        & \cellcolor[HTML]{EFEFEF}244.09                            & \textbf{\cellcolor[HTML]{EFEFEF}1.00}                                 \\ \cline{4-11} 
                        &                          &                           & Greedy                     & \multicolumn{1}{r|}{\textbf{36.15}}   & 8.88                         & \multicolumn{1}{r|}{\textbf{0.14}}    & 0.03                         & \multicolumn{1}{r|}{1650.90}                 & 434.67                            & \textbf{1.00}                                 \\ \cline{4-11} 
                        &                          &                           & \cellcolor[HTML]{EFEFEF}SMART                      & \multicolumn{1}{r|}{\cellcolor[HTML]{EFEFEF}300.22}           & \cellcolor[HTML]{EFEFEF}145.21                       & \multicolumn{1}{r|}{\cellcolor[HTML]{EFEFEF}2.16}             & \cellcolor[HTML]{EFEFEF}1.03                         & \multicolumn{1}{r|}{\cellcolor[HTML]{EFEFEF}3415.67}                 & \cellcolor[HTML]{EFEFEF}1358.18                           & \cellcolor[HTML]{EFEFEF}0.90                                \\ \cline{2-11} 
                        & \multirow{2}{*}{8}       & Cross                     & Greedy                     & \multicolumn{1}{r|}{\textbf{568.03}}  & 108.08                       & \multicolumn{1}{r|}{\textbf{52.73}}   & 98.85                        & \multicolumn{1}{r|}{\textbf{5899.60}}        & 572.84                            & \textbf{1.00}                                 \\ \cline{3-11} 
                        &                          & Random                    & \cellcolor[HTML]{EFEFEF}Greedy                     & \multicolumn{1}{r|}{\textbf{\cellcolor[HTML]{EFEFEF}611.60}}  & \cellcolor[HTML]{EFEFEF}166.98                       & \multicolumn{1}{r|}{\textbf{\cellcolor[HTML]{EFEFEF}9.56}}    & \cellcolor[HTML]{EFEFEF}9.07                         & \multicolumn{1}{r|}{\textbf{\cellcolor[HTML]{EFEFEF}2857.50}}        & \cellcolor[HTML]{EFEFEF}674.12                            & \textbf{\cellcolor[HTML]{EFEFEF}1.00}                                 \\ \cline{2-11} 
                        & \multirow{2}{*}{12}      & Cross                     & Greedy                     & \multicolumn{1}{r|}{\textbf{1701.46}} & 416.15                       & \multicolumn{1}{r|}{\textbf{571.01}}  & 1157.34                      & \multicolumn{1}{r|}{\textbf{9430.40}}        & 766.14                            & \textbf{1.00}                                 \\ \cline{3-11} 
                        &                          & Random                    & \cellcolor[HTML]{EFEFEF}Greedy                     & \multicolumn{1}{r|}{\textbf{\cellcolor[HTML]{EFEFEF}1644.71}} & \cellcolor[HTML]{EFEFEF}368.27                       & \multicolumn{1}{r|}{\textbf{\cellcolor[HTML]{EFEFEF}153.60}}  & \cellcolor[HTML]{EFEFEF}347.90                       & \multicolumn{1}{r|}{\textbf{\cellcolor[HTML]{EFEFEF}5383.10}}        & \cellcolor[HTML]{EFEFEF}924.73                            & \textbf{\cellcolor[HTML]{EFEFEF}1.00}                                 \\ \cline{2-11} 
                        & \multirow{2}{*}{16}      & Cross                     & Greedy                     & \multicolumn{1}{r|}{\textbf{3691.00}} & 580.97                       & \multicolumn{1}{r|}{\textbf{314.17}}  & 245.12                       & \multicolumn{1}{r|}{\textbf{12285.10}}       & 1158.18                           & \textbf{1.00}                                 \\ \cline{3-11} 
                        &                          & Random                    & \cellcolor[HTML]{EFEFEF}Greedy                     & \multicolumn{1}{r|}{\textbf{\cellcolor[HTML]{EFEFEF}3799.55}} & \cellcolor[HTML]{EFEFEF}844.19                       & \multicolumn{1}{r|}{\textbf{\cellcolor[HTML]{EFEFEF}174.95}}  & \cellcolor[HTML]{EFEFEF}135.29                       & \multicolumn{1}{r|}{\textbf{\cellcolor[HTML]{EFEFEF}7537.70}}        & \cellcolor[HTML]{EFEFEF}1509.79                           & \textbf{\cellcolor[HTML]{EFEFEF}1.00}                                 \\ \hline
\end{tabular}
\label{tab:sorting_results}
\end{table*}

The initial configuration of the objects is randomly generated in the common workspace.
The target configuration of the objects is randomly generated within the bin for the appropriate class.
The comparison of our approach to SMART~\cite{sb-smrgbmgwcc-20} is given in Figure~\ref{fig:sorting_results} and Table~\ref{tab:sorting_results}.

\subsubsection{Two Robot Scenario}
In this scenario, we consider two manipulators which must sort the blocks by color (Fig.~\ref{fig:physical_image}).
We evaluate the runtime and 
\textcolor{black}{solution quality} 
of the methods as the number of objects increases.

\subsubsection{Four Robot Scenario}
In this virtual scenario, a group of four robots again must sort an increasing number of blocks. 
This is an expansion of the Two Robot Scenario with the robots arranged in a square.
There are four classes of objects, one for each robot.
This evaluates the performance of the method as the number of robots increases and the impact the number of robots has on the performance of the method as the number of objects increases as well.

\subsubsection{Analysis}
\textcolor{black}{
As seen in Fig.~\ref{fig:sorting_results}, the DaSH variants are able to both find plans faster and for a significantly larger number of objects (20) than the SMART baseline (6).
SMART is unable to scale as the composite space it searches grows too large.
}

\textcolor{black}{
Both DaSH variants are able to efficiently find plans for smaller numbers of objects, and the heuristic based on the motion hypergraph effectively informs both variants on which action to take next.
MM-DaSH struggles with the task space complexity at six objects.
This occurs when there are a many roughly equivalent solution options with incompatible transition histories (e.g. robot 1 starting with block 1 or 2 when both need to be handed off to robot 2).
Greedy-MM-DaSH alleviates this issue by building a single transition history and is able to plan for
twenty objects for the two robot scenario and sixteen objects for the four robot scenario.
SMART is only able to obtain a 70\% success on the Random Scenario for two robots, six objects and a 0\% on the same Cross Scenario.
}

\textcolor{black}{
MM-DaSH produces lower cost solutions than either method though the Greedy-MM-DaSH solution cost is often close (Table~\ref{tab:sorting_results}).
SMART produces much higher cost solutions despite its much denser representation (using PRM* roadmaps).
This is due both to the requirement that all actions be performed synchronously and quality of the initial motion paths found using the dRRT* techniques (initial paths are used as we consider the initial solution produced by SMART).
In Table~\ref{tab:sorting_results}, we can see the construction time for SMART is an order of magnitude greater than the DaSH variants.
}

\textcolor{black}{
The Cross Scenario presents a more difficult problem for all methods as local decisions regarding individual objects often conflict with the best global decision regarding which object should be passed across the formation first.
This shows up in the hyperpath query stage for the DaSH variants and in the MAPF heuristic for SMART.
The variance is much higher for MM-DaSH and SMART in the Random Scenario as the more the problem resembles the Cross Scenario, the higher the difficulty and increased planning time.
Greedy-MM-DaSH does not exhibit this as it greedily selects an object to pass all the way to its goal in either scenario, ignoring the best global decisions.
In either scenario, most of the MM-DaSH and Greedy-MM-DaSH variance comes from the conflict resolution stage and is a product of the sampled transitions and roadmaps available to resolve conflicts.
}

\subsection{Shelving}
\begin{figure}[t]
    \centering
     \subfloat[Shelves - Scenario]{\includegraphics[width=.6\linewidth]{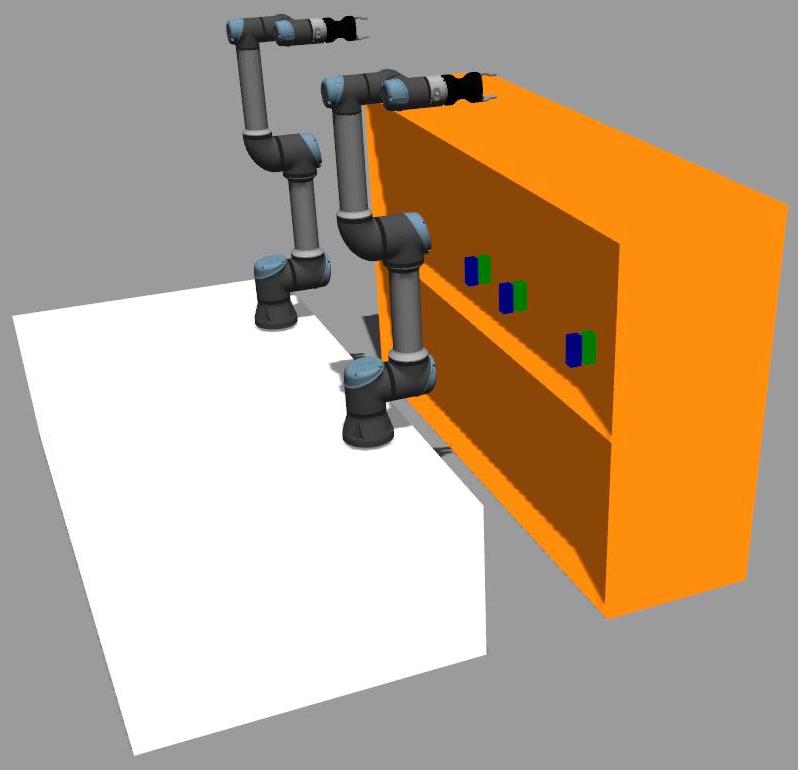}
    \label{fig:shelves_image}
    }\\
     \subfloat[Shelves - Results]{
    	\includegraphics[height=.6\linewidth]{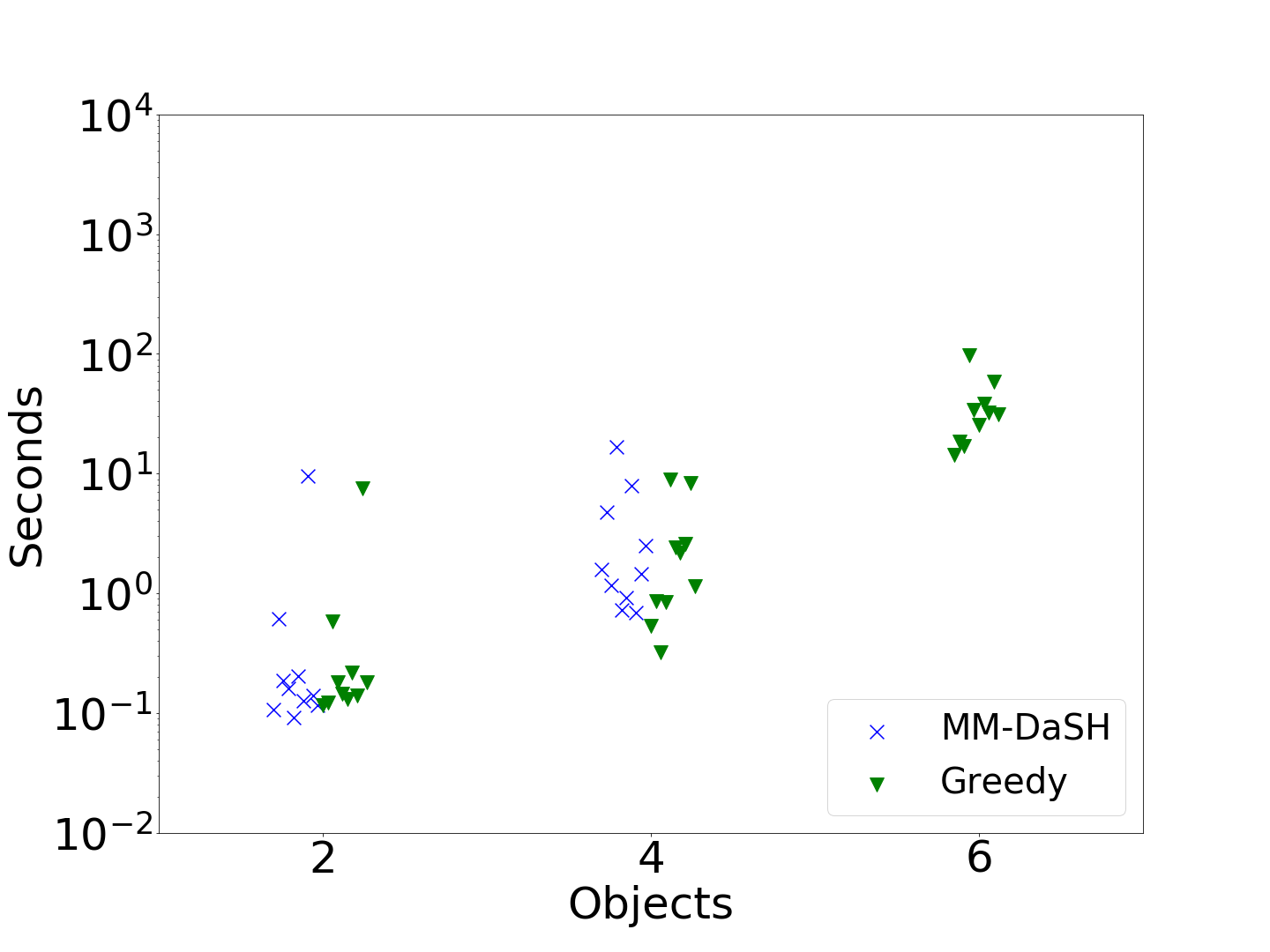}
    	\label{fig:shelves}
    }
    \caption{
    \textcolor{black}{
    (a) The robots must move the objects (which start randomly on the table) to positions on the shelf. The blue objects must be placed directly in front of the green objects, which requires the green objects to be placed first.
    (b) This figure reports the search time for each seed for the shelving scenario.
    The Y-axis reports seconds taken to find a solution.
    The X-axis reports the number of objects to be stocked.
    Both methods reported 100\% success for the 2 and 4 object problems.
    Greedy reported 100\% success for 6 objects.
    }
    }
    \label{fig:additional_results}
\end{figure}

\textcolor{black}{
To evaluate the ability of the method to account for geometrically constrained manipulation problems, we consider a shelf stocking scenario with two rows of objects (Fig.~\ref{fig:shelves_image}).
In this scenario, two manipulators must stock a shelf 
where the target configuration of objects on the shelf is random placements of object pairs such that one object lies immediately behind the other on the shelf, completely blocked by the front object.
This forces the method to reason over the collision-based scheduling constraints and determine the correct order to place the objects on the shelf.
The initial configuration of the objects is randomly generated on the same flat surface the robots rest on.
The results are shown in Fig.~\ref{fig:shelves}.
}

\subsubsection{Analysis}

\textcolor{black}{
The planning times shown in Fig.~\ref{fig:shelves} illustrate the ability of
both Greedy-MM-DaSH and MM-DaSH to efficiently find plans for this geometrically constrained problem, placing each green object in the back before placing its coupled blue object in the front (Fig.~\ref{fig:shelves})
with Greedy successfully finding plans for more objects than MM-DaSH.
They do take longer than the sorting scenarios as it often takes multiple iterations for the motion planning layer to identify each of the motion-based scheduling constraints. 
}


\section{Conclusion}
In this paper, we present the \textit{Decomposable State Space Hypergraph} (DaSH) method, a general task and motion planning framework which considers varying levels of coupled and decoupled spaces.
The framework enables tailoring of the space coupling to the coordination required for different problem stages rather than a static trade off between coordination and planning speed.
We provide a decomposition of task and motion planning spaces into subspaces and present a hypergraph representation capable of representing the decomposed space.
We present a method for leveraging this hypergraph representation to solve task and motion planning problems.

We illustrate the application of the general approach to the multi-robot motion planning problem (MRMP-DaSH) and the multi-manipulator problem (MM-DaSH).
We provide an analysis of the representation for the multi-manipulator problem and empirically evaluate our approach against a state-of-the-art multi-manipulator task and motion planning method where we show significant improvement in planning.


This improvement in planning is in large part due to the ability to construct the hypergraph-based representation.
This is not be the case for all task and motion planning problems.
Future work should consider how to reason over implicit representations of the decomposed space.

\section{Acknowledgements}
We would like to acknowledge the other members of the Parasol Lab at the University of Illinois Urbana-Champaign (UIUC) for their suggestions regarding this work. 
In particular, we are grateful to Isaac Ngui and Isaac Burton Love for their input and discussion and to Hannah Lee for the creation of the figures in this manuscript.
This work was performed with support from Foxconn Interconnect Technology (FIT) and the Center for Networked Intelligent Components and Environments (C-NICE) at UIUC.

\bibliographystyle{IEEEtran}
\bibliography{robotics.bib}



\end{document}